\PassOptionsToPackage{unicode}{hyperref}
\PassOptionsToPackage{hyphens}{url}
\PassOptionsToPackage{dvipsnames,svgnames,x11names}{xcolor}
\documentclass[
  10pt,
]{article}
\usepackage{xcolor}
\usepackage[margin=1in]{geometry}
\usepackage{amsmath,amssymb}
\usepackage{iftex}
\ifPDFTeX
  \usepackage[T1]{fontenc}
  \usepackage[utf8]{inputenc}
  \usepackage{textcomp} 
\else 
  \usepackage{unicode-math} 
  \defaultfontfeatures{Scale=MatchLowercase}
  \defaultfontfeatures[\rmfamily]{Ligatures=TeX,Scale=1}
\fi
\usepackage{lmodern}
\ifPDFTeX\else
\fi
\IfFileExists{upquote.sty}{\usepackage{upquote}}{}
\IfFileExists{microtype.sty}{
  \usepackage[]{microtype}
  \UseMicrotypeSet[protrusion]{basicmath} 
}{}
\makeatletter
\@ifundefined{KOMAClassName}{
  \IfFileExists{parskip.sty}{%
    \usepackage{parskip}
  }{
    \setlength{\parindent}{0pt}
    \setlength{\parskip}{6pt plus 2pt minus 1pt}}
}{
  \KOMAoptions{parskip=half}}
\makeatother
\usepackage{longtable,booktabs,array}
\usepackage{calc} 
\usepackage{etoolbox}
\makeatletter
\patchcmd\longtable{\par}{\if@noskipsec\mbox{}\fi\par}{}{}
\makeatother
\IfFileExists{footnotehyper.sty}{\usepackage{footnotehyper}}{\usepackage{footnote}}
\makesavenoteenv{longtable}
\usepackage{graphicx}
\makeatletter
\newsavebox\pandoc@box
\newcommand*\pandocbounded[1]{
  \sbox\pandoc@box{#1}%
  \Gscale@div\@tempa{\textheight}{\dimexpr\ht\pandoc@box+\dp\pandoc@box\relax}%
  \Gscale@div\@tempb{\linewidth}{\wd\pandoc@box}%
  \ifdim\@tempb\p@<\@tempa\p@\let\@tempa\@tempb\fi
  \ifdim\@tempa\p@<\p@\scalebox{\@tempa}{\usebox\pandoc@box}%
  \else\usebox{\pandoc@box}%
  \fi%
}
\def\fps@figure{htbp}
\makeatother
\providecommand{\tightlist}{%
  \setlength{\itemsep}{0pt}\setlength{\parskip}{0pt}}
\usepackage{bookmark}
\IfFileExists{xurl.sty}{\usepackage{xurl}}{} 
\makeatletter
\@ifundefined{xmpquote}{}{}
\makeatother
\hypersetup{
  colorlinks=true,
  linkcolor={black},
  filecolor={Maroon},
  citecolor={Blue},
  urlcolor={blue},
  pdfcreator={LaTeX via pandoc}}

\author{}
\date{}

\begin{document}

\section{Elicitation format changes zero-shot classification accuracy in
a medical vision-language model: a two-center evaluation on colorectal
histopathology}\label{elicitation-format-changes-zero-shot-classification-accuracy-in-a-medical-vision-language-model-a-two-center-evaluation-on-colorectal-histopathology}

\textbf{Running title.} Elicitation format changes the prediction, not
only the confidence.

\textbf{Authors.} Anay Sinhal, Department of Computer and Information
Science and Engineering and Department of Medicine, University of
Florida; Amit Sinhal, Department of Computer Science and Engineering, JK
Lakshmipat University; Arpana Sinhal, Department of Computer
Applications, Manipal University Jaipur. Correspondence: Arpana Sinhal,
arpana.sinhal@jaipur.manipal.edu.

\begin{center}\rule{0.5\linewidth}{0.5pt}\end{center}

\subsection{Abstract}\label{abstract}

\textbf{Background.} Verbalized confidence from a hosted medical
vision-language model is usually treated as packaging around an
unchanged prediction.

\textbf{Methods.} We held a nine-class colorectal histopathology prompt
fixed except for the requested response format and evaluated
MedGemma-27b-it on 10,004 validation and 7,180 external-center PathMNIST
patches. Endpoints were preregistered, ties randomized, and random
floors recomputed for each patch set.

\textbf{Results.} Requesting probabilities over all nine classes instead
of a ranked top-3 reduced accuracy by 14.8 points on validation, -0.1475
{[}-0.1594, -0.1353{]}, and 13.4 at the external center, -0.1341
{[}-0.1484, -0.1196{]}, against chance accuracy of 0.1111. The
nine-class prompt uniquely forbade commentary, reasoning, and a chosen
answer. Removing that prohibition at matched cardinality and token
budget recovered +0.1878 {[}+0.1623, +0.2134{]} and +0.1911 {[}+0.1648,
+0.2181{]}. Changing cardinality with the prohibition absent moved
accuracy by -0.0013 {[}-0.0300, +0.0280{]} and -0.0214 {[}-0.0494,
+0.0071{]}. The nine-class format returned no parseable distribution on
7.6 and 7.7 percent of patches; zero vectors accounted for 94.0 and 98.6
percent of those failures, concentrated in the two least-textured
classes. Deferring failures first explained its measurable
selective-prediction gain. On a featureless image, the prohibited prompt
refused every call, the ranked top-3 returned a well-formed prior, and
the nine-class request without the prohibition never returned a zero
vector. Across four vision-capable models, accuracy did not grade the
cost: two models matched on accuracy differed five- to sevenfold.
First-token logit entropy routed better than any elicited signal,
+0.1820 {[}+0.1722, +0.1974{]} versus +0.0493 {[}+0.0377, +0.0649{]}
relative to each signal's random floor.

\textbf{Conclusions.} Response format changes a medical vision-language
classifier. Here, the operative component was the instruction
suppressing an answer. Use logits when available; otherwise request a
short ranked list without prohibiting reasoning or a chosen answer.

\textbf{Keywords.} verbalized confidence, elicitation format, selective
prediction, medical vision-language models, computational pathology,
structured output.

\begin{center}\rule{0.5\linewidth}{0.5pt}\end{center}

\section{1. Introduction}\label{introduction}

A clinician-facing triage system built on a hosted medical
vision-language model can change the prompt and read the text back.
Often that is all it can do: no weights, no hidden states, and on many
commercial endpoints no logprobs either. In that setting the only
uncertainty signal available is the one the model can be persuaded to
say out loud, and the literature on verbalized confidence grew up around
exactly this constraint (Lin et al., 2022; Tian et al., 2023; Xiong et
al., 2024). It asks which elicited object is best calibrated, or which
best predicts an error. It does not ask what the act of asking does to
the answer.

We asked MedGemma-27b-it (Sellergren et al., 2025) to classify
colorectal histopathology patches into nine tissue types and varied only
the shape of the response requested. Every variant carries the same
preamble and the same per-class cue block; the variants differ in
whether they ask for a probability over all nine classes or for a ranked
top-3 with probabilities. Nothing about the image, the task, the
decoding, or the description of the classes changes between them.

One of those two requests destroys the classifier. As a paired
difference on the patches where both formats produced an answer, argmax
accuracy falls by -0.1475 {[}-0.1594, -0.1353{]} on the validation split
and -0.1341 {[}-0.1484, -0.1196{]} at an external clinical center,
taking a classifier at 0.3817 and 0.4216 to 0.2223 and 0.2720 against a
chance level of 0.1111. The three-number request itself, under the
identical cue block, costs 0.0264 against a single-label prompt, so
almost the whole of the cost lands in the step between those two prompts
rather than accumulating with output length.

Naming what in that step does it took a third prompt. The two requests
differ in two ways, not one: in how many classes they ask for, and in a
closing sentence carried only by the nine-class prompt, forbidding
commentary, reasoning and a chosen answer. A variant that makes the
nine-class demand without the prohibition separates them, and the
prohibition carries the whole of the cost while the cardinality carries
none we can detect.

Neither claim on its own would be new: that response format matters is
established, and so is a format demand costing accuracy. Tam et
al.~(2024) measured format restriction against task performance in text
models and found reasoning degraded while classification, on the whole,
did not; JSON-mode beat free text on all five models tested on a medical
differential-diagnosis benchmark. Taveekitworachai et al.~(2026) put
fifteen medical reasoning models under three requested answer formats
and found format robustness ranging from 35 to 100 percent, which is a
medical result on a correctness endpoint driven by the requested format.
Lee et al.~(2026) separate the format request from the decoder
constraint across ten models and locate most of the accuracy loss in the
instruction itself. So format-driven degradation of a classifier is a
text-domain result that already exists, including in medicine, and the
nearest prior measurement on a medical classification task runs opposite
to ours.

Two qualifiers carry what is left. No such result is reported in vision,
and none of this prior work studies a request for a \emph{distribution
over a closed label space} rather than a wrapper around a chosen label,
which is the intervention that fails here and the one a practitioner is
most likely to reach for when they want a confidence number.

Because the format changes the prediction, it also changes what can be
routed. We evaluate every elicited object as a selective-prediction
signal, in the framing that has become standard for deciding when a
medical model should defer to a human (Chow, 1970; Leibig et al., 2017;
Dolezal et al., 2022): does the signal order a fixed model's errors well
enough to be worth deferring on? Three results follow. The nine-class
format's one piece of routing value is its own failure, since a rule
that defers unparseable patches first accounts for the whole of its
measurable selective-accuracy gain and the numbers it does emit add
nothing detectable on top. The ranked top-3 format does route, and beats
the single-query stated confidence it replaces at both centers with the
cue block held fixed. And the elicited object loses to a signal that
arrives free in the same response: the normalized entropy of the top-5
logits at the first content-bearing token clears its own recomputed
random floor by +0.1820 {[}+0.1722, +0.1974{]} where the elicited object
manages +0.0493 {[}+0.0377, +0.0649{]}. That single comparison needs
logprob access; nothing else here does.

Four contributions follow, ordered by how far each travels beyond this
benchmark.

\begin{enumerate}
\def\labelenumi{\arabic{enumi}.}
\item
  \textbf{A decomposition that separates what a selective-prediction
  gain is made of.} Any method that cannot score some of its inputs
  earns area under an accuracy-coverage curve twice over: once by
  deferring the inputs it cannot score, which any method with the same
  coverage would earn, and once by ordering the inputs it can. We
  separate the two exactly. The separation is arithmetic and applies to
  any selective-prediction system with incomplete coverage.
\item
  \textbf{The format effect, measured, and the sentence that causes it.}
  A response-format change costs 13 to 15 argmax accuracy points in a
  medical vision-language model, at natural class prevalence on two
  whole splits from different clinical centers, with intervals, a
  preregistered control variant and a replication. A third variant
  locates the cause in the prohibition rather than the cardinality. The
  instruction one writes to get parseable output is what destroys the
  classifier, and the number of classes requested costs coverage rather
  than accuracy.
\item
  \textbf{What the failure is, and what triggers it.} The characteristic
  failure is not a malformed answer but a vector of nine zeros, the
  model naming every class in the requested order and assigning none of
  them any probability. A blind-image control reproduces it on a
  featureless field, where the ranked top-3 format returns a well-formed
  prior, locating the trigger in featureless rather than uninformative
  input.
\item
  \textbf{Four things the effect is not, each excluded by a
  measurement.} Not model headroom, since across four vision-capable
  models the cost is not a smooth function of task accuracy and two
  models matched on accuracy differ in cost by a factor of five to
  seven. Not the decoder, which we never constrain and cannot. Not a
  parse artefact, since the effect grows when unparseable responses are
  counted as errors. And not a scoring convention, since two ways of
  extracting a label disagree on 1.9 percent of responses at natural
  prevalence and we report both throughout.
\end{enumerate}

Section 3.11 says which analyses were confirmatory and which were not.
The code, the prompts, the preregistration and the raw per-patch model
responses behind every number are released as described under Code
availability.

\begin{center}\rule{0.5\linewidth}{0.5pt}\end{center}

\section{2. Related work}\label{related-work}

Three bodies of work bear on this question and they answer different
halves of it. The format literature establishes that how a request is
worded changes what a model returns, almost entirely in text. The
confidence-elicitation literature establishes what a stated confidence
is worth, almost entirely where the stated confidence already works. The
medical imaging literature establishes what a deferral system has to
clear to be useful. We take ground in the first and concede it in the
other two.

\subsection{2.1 What a response format does to the
answer}\label{what-a-response-format-does-to-the-answer}

That output format is not neutral is established. Sclar et al.~(2024)
showed that semantically equivalent formatting choices move few-shot
accuracy across a range wide enough to reorder models, which made format
a variable to control rather than a detail to fix arbitrarily. Long et
al. (2025) measured the same axis across fifteen formats and eight tasks
and found the variance attributable to the wrapping format alone
reaching 235.33 for a single model, reading the effect largely as a
failure to follow the format instruction.

Where that literature disagrees with itself is on whether the cost falls
on reasoning or on classification, and that is where our result enters.
Tam et al.~(2024) is the closest prior measurement and it points the
other way from ours: comparing free-form generation against JSON-mode
and format-restricting instructions across five models, they found
reasoning degraded monotonically in the strictness of the constraint
while classification, on the whole, improved, and the set that improves
most cleanly is a medical differential-diagnosis benchmark. The nearest
prior result to ours is therefore a format restriction substantially
helping a medical classification task, against a 13 to 15 point collapse
on a medical vision one.

Both can be true, and the reconciling observation is in what is being
restricted. Tam et al. constrain the \emph{syntax} around a single
label. The format that fails here changes the \emph{cardinality of what
must be produced}, from one label to nine numbers that have to sum to
one. Section 4.1 controls for that distinction directly.

The direction we report is not new in text. Taveekitworachai et
al.~(2026, preprint) pose the same medical questions to fifteen
proprietary and open-weight medical reasoning models under three
requested formats and measure correctness rather than agreement;
answer-format robustness varies from 35 to 100 percent across the
fifteen. That is a medical result, on a correctness endpoint, driven by
the requested format, and it precedes us. It is in text, on question
answering, and its formats wrap an answer rather than requesting a
distribution over a fixed label space, so it does not reach the
intervention we run, but it retires the word \emph{medicine} from any
claim of priority.

Lee et al.~(2026, preprint) anticipate our mechanism argument in a
weaker form. Across six open-weight and four API models, four formats
and tasks spanning mathematics, science, logic and writing, they locate
the dominant cost of structured output at the prompt: format-requesting
instructions alone account for most of the accuracy loss \emph{before}
any decoder constraint is applied, and constrained decoding's sampling
bias explains only a fraction. Section 5.1 reaches the same conclusion
from an endpoint that returns no logits and cannot constrain decoding at
all.

One account runs directly against this framing. Fan (2026, preprint)
argues that degradation attributed to format is a capacity effect:
models with spare headroom absorb a structural constraint at no cost,
88.7 percent against 89.3 on MATH-Hard for the strongest model tested,
while models near their limits degrade through truncation under a
standard token budget, worth 36.2 points on one model, and through
capacity competition that survives eliminating truncation, worth 28.0 on
another. Under that reading the operative variable is headroom and
format is only what consumes it, so the same demand would be free on a
stronger model. Sections 4.6 and 5.3 take it up with a measurement.

What is left after all of this is narrow. A response format degrading a
classifier is reported in text, by Tam et al.'s own results, by Lee et
al.~generally and by Taveekitworachai et al.~in medicine. What is not
reported anywhere our searches reached is such a result in vision, or
for a requested \emph{distribution over a closed label space} rather
than a wrapper around a chosen label.

\subsection{2.2 What a stated confidence is
worth}\label{what-a-stated-confidence-is-worth}

Verbalized confidence became a research object because hosted endpoints
stopped returning anything else, and because a model's stated
self-assessment carries real information: Kadavath et al. (2022,
preprint) showed that a model asked whether its own answer is correct
answers better than chance across a wide task range, which is the
premise every elicitation method here inherits. Lin et al.~(2022) taught
a model to state calibrated uncertainty in words and compared that
channel directly against logit-derived uncertainty, which is the
comparison we run in section 4.4. Tian et al.~(2023) established that
for models fine-tuned with human feedback the verbalized channel often
beats the model's own conditional probabilities on factual question
answering, and that asking for several candidate answers before a
confidence improves the verbalized score.

What we evaluate is an elicited ranked top-k, which is Tian's
second-best strategy, and we do not claim it as new. The regime is what
differs. Tian's models are miscalibrated; the single-query stated
confidence on this task is at chance, with an area under the receiver
operating characteristic curve (AUROC) against per-patch correctness of
0.4975 {[}0.4938, 0.5012{]}. ``Better calibrated'' and ``recovers a
usable ordering from a signal at chance'' are different claims and only
the second is at issue here.

Two entries argue against us and are read that way. Xiong et al.~(2024)
find a hybrid of elicitation and consistency beating either alone in 13
of 20 cases, which points toward combining a single-query object with a
multi-query one rather than replacing it; our comparator results agree,
since against a five-query baseline we are not distinguishable at either
center. Groot and Valdenegro-Toro (2024) is the earliest record of
verbalized uncertainty elicited from a vision-language model by
prompting, and the correct citation for the fact that such confidence is
overconfident; their framing is also the cleanest argument for reporting
a ranking metric rather than a calibration one, since a signal can be
uniformly overconfident and still rank correctly.

The baseline an elicited object is meant to displace is multi-query
agreement. Aggarwal et al. (2023) show adaptive stopping recovers much
of its cost, so our cost ratio is reported under a stopping assumption
as well as at full K. That baseline is a paraphrase-agreement score
rather than self-consistency, for the reason given in section 3.7.

\subsection{2.3 Deferral in medical imaging, and the ceiling it has to
clear}\label{deferral-in-medical-imaging-and-the-ceiling-it-has-to-clear}

This is the literature that decides whether any of this matters
clinically. Its long-standing position is that a model's usefulness in a
clinical workflow depends on whether it can say when not to be trusted
(Begoli et al., 2019; Kompa et al., 2021), and that the operational form
of that is a confidence threshold above which the model answers and
below which a human does. Leibig et al.~(2017) demonstrated the pattern
on diabetic retinopathy, where referring the least confident cases
raised accuracy on the retained ones enough to change the deployment
case. Dolezal et al.~(2022) is the direct precedent in our own domain:
confidence thresholding on digital histopathology, evaluated across
institutions. The endpoint this paper measures is theirs; what we vary,
and they do not, is how the confidence is obtained.

The framing has a formal core older than the deep-learning literature
that adopted it. Chow (1970) derived the error-rejection tradeoff for a
classifier permitted to abstain. El-Yaniv and Wiener (2010) and Geifman
and El-Yaniv (2017) recast it as selective classification, and Kamath et
al. (2020) extended it to distribution shift, which is the regime a
second clinical center puts us in. Traub et al.~(2024) argue that common
multi-threshold metrics in this family fail five requirements they
define and propose an alternative that changes metric rankings on five
of six benchmarks; we do not adopt it, and section 6.4 gives the reason
and what it costs.

Deep learning on histopathology has moved from task-specific classifiers
(Kather et al., 2019) to slide-level and vision-language foundation
models (Huang et al., 2023; Lu et al., 2024; Xu et al., 2024), with the
clinical-translation literature consistently identifying reliability
rather than accuracy as the binding constraint (Niazi et al., 2019; van
der Laak et al., 2021). MedGemma (Sellergren et al., 2025, preprint)
belongs to the generalist-medical class that literature anticipated
(Moor et al., 2023). Nothing in that line asks what the elicitation
format does to the prediction, because the models it studies expose
their own probabilities and nobody has to ask.

\subsection{2.4 The closest studies,
compared}\label{the-closest-studies-compared}

\textbf{Table 1. The studies this work is positioned against.}
\texttt{PR} marks a peer-reviewed version of record; \texttt{PP} marks a
preprint with none at the time of writing.

{\def\LTcaptype{none} 
\begin{longtable}[]{@{}
  >{\raggedright\arraybackslash}p{(\linewidth - 6\tabcolsep) * \real{0.2500}}
  >{\raggedright\arraybackslash}p{(\linewidth - 6\tabcolsep) * \real{0.2500}}
  >{\raggedright\arraybackslash}p{(\linewidth - 6\tabcolsep) * \real{0.2500}}
  >{\raggedright\arraybackslash}p{(\linewidth - 6\tabcolsep) * \real{0.2500}}@{}}
\toprule\noalign{}
\begin{minipage}[b]{\linewidth}\raggedright
Study
\end{minipage} & \begin{minipage}[b]{\linewidth}\raggedright
Manipulation and setting
\end{minipage} & \begin{minipage}[b]{\linewidth}\raggedright
Key finding
\end{minipage} & \begin{minipage}[b]{\linewidth}\raggedright
Limitation for our question
\end{minipage} \\
\midrule\noalign{}
\endhead
\bottomrule\noalign{}
\endlastfoot
\textbf{Format as an intervention on the prediction} & & & \\
Tam et al.~(2024) \texttt{PR} & Free text vs.~format-restricting
instruction vs.~JSON-mode, 5 LLMs; text reasoning and classification
incl.~medical differential diagnosis & Reasoning degrades monotonically
in strictness; classification largely improves, by 11.3 to 30.1 points
on the medical set & Text only; restricts syntax around one label, never
the cardinality of the output; no vision \\
Sclar et al.~(2024) \texttt{PR} & Semantically equivalent formatting
perturbations, few-shot text & Format alone moves accuracy enough to
reorder models & Establishes format as a variable, not what it costs a
classifier \\
Long et al.~(2025) \texttt{PR} & 15 wrapping formats, 8 tasks, text &
Cross-format variance of one model reaching 235.33; read as
instruction-following failure & Text; no distribution over a closed
label space \\
Lee et al.~(2026) \texttt{PP} & Format request vs.~constrained decoding,
separated; 6 open-weight and 4 API models, 4 formats; text & The
instruction alone causes most of the loss, before any decoder constraint
& Text; our endpoint exposes no decoder to constrain, so we confirm
their prompt-side half only \\
Taveekitworachai et al.~(2026) \texttt{PP} & Multiple-choice
vs.~open-ended vs.~ranked list, 15 medical reasoning models; text QA &
Answer-format robustness ranges from 35 to 100 percent & Text and
question answering; formats wrap a chosen answer, never request a
distribution; no vision \\
Fan (2026) \texttt{PP} & Structured vs.~free output across models of
differing headroom, token budget varied; text mathematics & Strong
models absorb structure at no cost; weak ones lose to truncation and
capacity competition & Predicts our result; section 4.6 tests it
directly \\
\textbf{What an elicited confidence is worth} & & & \\
Tian et al.~(2023) \texttt{PR} & Verbalized confidence, several
strategies incl.~top-k candidates; RLHF-tuned models, text factual QA &
Verbalized often beats the model's own conditional probabilities; top-k
improves it further & Regime where the scalar self-report already works;
calibration endpoint, not ranking \\
Xiong et al.~(2024) \texttt{PR} & Elicitation, sampling and aggregation
benchmarked on calibration and failure prediction; text & Consistency
beats vanilla verbalized; hybrids best in 13 of 20 cases & Argues for
combining elicitation with agreement, not replacing it; text only \\
\textbf{Deferral in medical imaging} & & & \\
Dolezal et al.~(2022) \texttt{PR} & Uncertainty-thresholded deferral,
within and across institutions; digital histopathology & High-confidence
subsets are substantially more accurate; the remainder is deferred &
Confidence comes from the model's own probabilities, so the format
question does not arise \\
\end{longtable}
}

Two things follow from the table. Every row that manipulates format
works in text, and the row that works in medical imaging holds the
format fixed. The intersection is empty, and that intersection is what
section 4 reports.

\begin{center}\rule{0.5\linewidth}{0.5pt}\end{center}

\section{3. Methods}\label{methods}

Figure 1 gives the inference pipeline and Figure 2 the analysis
workflow. Prompts, provenance, the audit trail and two corrections to
our own machinery are in Supplementary Material S2.

\subsection{3.1 Task, data, and what ``cross-center'' is worth
here}\label{task-data-and-what-cross-center-is-worth-here}

We evaluate nine-class colorectal tissue classification on PathMNIST
from MedMNIST v2 (Yang et al., 2023), using the 224-pixel release. The
classes are adipose, background, debris, lymphocytes, mucus, smooth
muscle, normal colon mucosa, cancer-associated stroma, and colorectal
adenocarcinoma epithelium. Validation is the 10,004-patch held-out tenth
of NCT-CRC-HE-100K; the external test set is CRC-VAL-HE-7K, 7,180
patches from a different clinical site.

The phrase ``cross-center'' needs qualifying. NCT-CRC-HE-100K is itself
drawn from 86 slides at two institutions (Kather et al., 2018), so this
is one site against a two-site training distribution rather than one
site against another, and the patches are Macenko colour-normalised at
source. Both facts make the shift weaker than ``external validation''
usually implies, and bound what the cross-center instability of section
4.5 can establish about robustness.

\subsection{3.2 Model, serving, and
decoding}\label{model-serving-and-decoding}

The primary model is MedGemma-27b-it (Sellergren et al., 2025), a
medical vision-language model built on Gemma 3 (Gemma Team, 2025).
Section 3.10 adds three further models.

\textbf{Serving.} All inference ran against the University of Florida
NaviGator AI service, an OpenAI-compatible HTTP endpoint operated by UF
Research Computing. Requests reach a LiteLLM gateway, version 1.88.5,
which routes to an OpenAI-compatible inference server on an NVIDIA
B200-class GPU node serving the published
\texttt{google/medgemma-27b-it} checkpoint at a 128K context window.
Per-key throughput is capped at 120 requests per minute; we ran at a
concurrency of 8.

Every number here was produced through that one serving path, which is
fixed within this study rather than established across deployments. On a
hosted endpoint the raw responses are the only reproducibility guarantee
available, so every response is released. Section 6.1 states what a
second serving path would settle.

\textbf{Decoding.} Temperature 0.0, \texttt{top\_p} and \texttt{seed}
unset so endpoint defaults apply, and a 256-token completion cap, held
identical across every variant and every model so that the requested
response format is the only thing differing between the runs the
headline compares. Section 3.10 reports a second temperature as a
declared check. The endpoint is close to deterministic under this
configuration: across the 1,800 calls of a run holding the image
byte-identical, the nine-class format returned one distinct response and
the ranked top-3 format returned two, the minority form appearing 19
times.

\subsection{3.3 Elicitation variants}\label{elicitation-variants}

We compare six free-text elicitation formats differing in the response
format requested. Four ask for a probability over all nine classes (D1,
D3, D4, D6) and two ask for a ranked top-3 with probabilities (D2, D5).
Crossed with that, three carry the per-class cue block (D4, D5, D6) and
three do not (D1, D2, D3), which isolates the cue block's contribution
from the format's.

The two the paper turns on are D5, a ranked top-3, and D6, a nine-class
distribution, and they are not identical apart from the format sentence.
They differ in two ways. The first is the requested cardinality, three
ranked classes against nine. The second is that D6 closes with
\emph{``Do not add commentary, reasoning, a chosen answer, or any text
before or after the nine lines''} while D5 closes with \emph{``Respond
in exactly this format and nothing else''} and forbids nothing. D6
therefore suppresses a chosen answer explicitly, while D5's rank-1 entry
is a chosen answer, so the D5-minus-D6 argmax difference on its own
cannot separate cardinality from suppression. D4 is that separation: the
same preamble, the same cue block and the same nine-class demand,
closing on D5's wording with no prohibition. Section 3.12 reports the
control run that uses it.

One naming convention holds throughout. Two variants ask for nine
probabilities, so \emph{the nine-class format} without qualification
means D6, the variant of the headline pair, which asks for nine numbers
\emph{and} forbids commentary, reasoning and a chosen answer; a claim
about the nine-class demand independent of that sentence names D4. All
six variants are reproduced in Supplementary Material S2, and prompt
files are immutable once a run has used them.

The primary endpoint changed after the first data existed. The
preregistered primary endpoint was Shannon entropy over a self-reported
nine-class distribution. On a 100-patch pilot every nine-class variant
collapsed the classifier, argmax accuracy falling to chance while the
same model given the same cue block and asked for a ranked top-3 did not
collapse. The primary endpoint moved to entropy over the elicited ranked
top-3, the move was recorded before any scored run existed, and the
nine-class variant stayed in the confirmatory family as a control. Two
things follow: the amended endpoint is a top-k object close to existing
precedent (Tian et al., 2023); and every top-3 result was selected, in
part, by a measurement made on 100 patches.

\subsection{3.4 Parsing, and the criterion stated separately for each
format}\label{parsing-and-the-criterion-stated-separately-for-each-format}

A response is parsed into a probability vector over the nine classes.
The two formats are subject to different criteria by construction, and
both are stated because the coverage comparison between them turns on
where the rules differ.

\begin{itemize}
\tightlist
\item
  \emph{Nine-class formats (D1, D3, D4, D6).} All nine classes named
  exactly once, each with a stated probability, and the stated
  probabilities summing to within 0.02 of one. A sum outside tolerance
  is a failure and is never renormalised into existence. D4 is scored
  under this rule, which makes its coverage in section 4.8 comparable to
  D6's.
\item
  \emph{Ranked top-3 formats (D2, D5).} Three distinct named classes
  with stated probabilities, summing to no more than 1.02. A sum below
  one is not a failure: the prompt asks for the three most likely
  classes, so the shortfall is what the response asserts about the rest
  and is carried as an explicit fourth outcome. Over-unity is a failure.
\end{itemize}

The residual mass is treated as an outcome rather than as rounding
error, which is the operative definition of the primary endpoint. When a
top-3 response assigns 0.6, 0.2 and 0.1, the missing 0.1 enters the
entropy as a fourth outcome; renormalising would assert a confidence the
model did not express, precisely on the low-confidence patches a routing
signal exists to find. The renormalised variant is reported as a
declared sensitivity.

Failed responses are never renormalised, never imputed, and never
retried. They are counted, classified by failure mode, and reported as a
coverage cost. Coverage for the primary variant was 9,927 of 10,004 on
validation and 7,153 of 7,180 at the external center, and failures are
not uniform across classes: the primary variant parses 0.9708 {[}0.9575,
0.9800{]} of mucus against 0.9923 {[}0.9904, 0.9938{]} overall, and
mucus is a class the model almost never classifies correctly, so parse
failure and accuracy failure coincide.

Because the two criteria differ, section 4.2 reports an equal-footing
check: every nine-class vector failing the sum tolerance is
renormalised, exactly the licence the top-3 format is given, and the
comparison is recomputed.

\subsection{3.5 The class-stratified scoping
rule}\label{the-class-stratified-scoping-rule}

Every quantity is reported on the whole split where one exists. Two of
the three evaluation splits available to us are class-balanced samples
of 1,800 patches, 200 per class, drawn from splits whose natural class
frequencies differ from uniform by up to a factor of four. A quantity
measured on such a sample is a property of the sample until it is
checked.

The largest distortion we found is the nine-class control's
selective-accuracy area at the external center: 0.1828 on the
class-balanced 1,800 against 0.3184 on the remaining 4,921 patches of
the same split, a difference of -0.1356 {[}-0.1672, -0.1011{]} from an
unpaired bootstrap on two disjoint sets. The balanced sample
over-represents exactly the classes the format fails on, so it
\emph{understated} a collapse rather than manufacturing one.

The rule we adopted: no quantity measured on the class-balanced subsets
is described as a property of the model, of the center, or of the
elicitation format until it has been checked at natural prevalence, and
every table says which basis a number is on. One conclusion of ours
reverses under it, and section 4.3 reports it. Section 6.3 lists what
remains on the subset basis and section 3.8 explains why spending cannot
fix it.

\subsection{3.6 Selective prediction, the floor, the decomposition, and
the tie
protocol}\label{selective-prediction-the-floor-the-decomposition-and-the-tie-protocol}

The routing question is whether a confidence score orders a fixed
model's errors well enough to be worth deferring on (Chow, 1970;
El-Yaniv and Wiener, 2010; Geifman and El-Yaniv, 2017). We measure it
with the area under the accuracy-coverage curve (AAUC). Three details of
the protocol matter more than they usually would, because our scores are
coarse.

Ties are broken uniformly at random within each block and averaged over
S = 200 seeds, with a percentile interval across seeds. A self-reported
confidence produces large tie blocks; the single-query scalar takes four
distinct values with a tie fraction above 0.91, and resolving ties by
array order lets the sort inherit whatever structure the patch ordering
carries. S was raised from 30 after a sensitivity analysis showed the
mean settling by S = 20 and the interval not until S = 100.

The random floor is recomputed on every patch set. A random ordering of
a fixed patch set has an AAUC equal to that set's overall accuracy, so
the floor is a property of the patches and of the classifier being
routed rather than a constant. Two signals attached to two different
predictions have two different floors, and comparing their raw AAUC
values compares two things at once. That matters here because the two
formats produce \emph{different classifiers}.

We therefore decompose. Writing AAUC = floor + (ceiling minus floor)
times ranking efficiency, where the ceiling is the analytic best
attainable at that accuracy and sample size, separates how accurate a
system is from how well it orders its own errors. Both components are
reported, and the decomposition is exploratory (section 3.11).

How much of the ordering the tie-break decides is reported as its own
quantity, measured by scoring the same signal against itself under two
independent tie-break draws on one shared patch resample. For the
nine-class control at full scale it is 0.0099 and 0.0124, more than half
of what those designs could detect; for the primary variant it is 0.0040
and 0.0042. Neither the distinct-value count nor the tie fraction
predicts it, and the quantity is reported beside every null it
qualifies.

\subsection{3.7 Comparators, and two label-extraction
conventions}\label{comparators-and-two-label-extraction-conventions}

The elicited object is compared against three baselines: the
single-query scalar confidence the same model emits under an unmodified
zero-shot prompt, an agreement score over five responses, and the
predictive entropy of the five labels. The five-query baseline is not
self-consistency and should not be cited as such: its five queries run
at temperature 0.0 with the prompt wording varied between them, so it
measures sensitivity to paraphrase rather than sampling variance
(contrast Wang et al., 2023).

A zero-shot vision-language classifier does not return a label; it
returns text from which one is extracted, and the extraction rule is a
methodological choice that has not been treated as one. Two conventions
appear here and we report both. \textbf{Sweep-based extraction}
lowercases the whole response and returns the first of nine ordered
regular expressions that matches anywhere in it, so a response whose
label field reads \texttt{LABEL:\ background} and whose free text
happens to contain ``debris'' is recorded as \texttt{debris}.
\textbf{Stated-label extraction} reads the model's own \texttt{LABEL:}
line.

The two agree on 98.07 percent of full-split validation responses and
98.29 percent at the external center, and the disagreements are
structural rather than scattered: 393 of the 515 validation confusions
are \texttt{background} read as \texttt{debris}. The rate is itself
class-dependent, 5.7 percent on the class-balanced subsets against 1.93
percent at natural prevalence, so a divergence rate is quoted here with
the basis it is on.

Stated-label extraction is primary and sweep-based extraction is
reported beside it in every affected table. The comparator traces
available to us were built under the sweep-based convention while the
elicited object was not, so comparing them under one convention compares
two things at once. Three verdicts depend on the choice, and section
4.4.1 reports them. Neither preregistered gate depends on the choice:
both are adjudicated on each signal's gap to its own recomputed floor,
and the primary variant is byte-identical under either convention,
+0.0880 {[}+0.0514, +0.1142{]} and +0.1190 {[}+0.0942, +0.1580{]}. Token
accounting is untouched, because token counts come from the endpoint's
usage field and never pass through an extractor.

\subsection{3.8 The scale asymmetry, and why it is
permanent}\label{the-scale-asymmetry-and-why-it-is-permanent}

The comparison against the two five-query baselines is at n near 2,000
while the gap-to-floor results are at full scale, and that gap cannot be
closed by buying more inference. The five-query traces exist only on the
1,800-patch class-stratified subsets; producing a full-scale version
would cost 85,920 calls.

We extended those subsets once, by 205 validation and 339 test patches,
under a selection rule fixed before the draw, sized to bring the minimum
detectable effect to the preregistered 0.02 practical threshold. It did
not get there and could not have. The realised value on the pooled
subset plus extension under stated-label extraction is 0.0517 at n =
1,981 on validation and 0.0526 at n = 2,124 at the external center, so
3,264 calls bought about 0.003 of resolution against a shortfall above
0.03. No further patches were drawn, because a stopping rule that reacts
to the realised resolution is outcome-dependent.

The threshold is unreachable on this dataset at any budget. The minimum
detectable effect scales as the inverse square root of the sample size,
verified by subsampling over thirty-two rungs with three draws each,
every rung landing between 0.962 and 1.027 times the prediction. The
whole validation split would deliver 0.0230 and the whole external test
split 0.0286; reaching 0.02 needs roughly 13,200 and 14,700 patches per
center and the splits contain 10,004 and 7,180. Pooling is foreclosed
because the claim is per-center. Every comparator interval containing
zero is therefore reported as not distinguishable at this sample size,
never as equivalence.

The subsets are also not representative of their splits, and in opposite
directions (section 6.3). The extension was drawn from the same
populations under the same rule, so it inherits that bias by
construction: it bought power, not validity.

Every minimum detectable effect here is quoted with its patch basis, its
n and its extraction convention, because the same design has five;
Supplementary Material S2 tabulates all five.

\subsection{3.9 The coverage treatment}\label{the-coverage-treatment}

A patch that cannot be scored cannot be routed. To bound what unscored
patches could hide, we re-run every comparison with each unscored patch
assigned the worst confidence rank and counted as an error, and
recompute the random floor on the enlarged denominator. This is the
standard selective-prediction treatment of a case a method cannot route:
the system does not get to pretend the patch is absent, and it is not
charged with confidently getting it wrong, because it never expressed a
confidence.

Both columns are reported for every signal, with an interval on their
difference, because a treatment applied only where it changes a verdict
is a treatment chosen for its verdict. For the primary variant, at 99.2
and 99.6 percent coverage, nothing moves: the change in its gap to floor
is +0.0025 {[}-0.0119, +0.0168{]} and +0.0011 {[}-0.0153, +0.0178{]} at
full scale. For the nine-class control the verdict changes at both
centers (section 4.2).

Two consequences govern how it is reported, both derived in
Supplementary Material S2: a difference between two coverage treatments
is a paired resample rather than a subtraction of the two levels,
because the levels sit on different denominators; and the gain the
treatment produces decomposes exactly into a deferral term and a score
term, which is what section 4.2 turns on.

\subsection{3.10 Controls, replication, and
sensitivity}\label{controls-replication-and-sensitivity}

Six checks bound the result. Section 3.11 says which are confirmatory.

\textbf{Granularity.} To separate information from resolution, we
quantize the elicited entropy by equal-frequency binning down to the
comparator's cardinality and re-run the comparison. The
matched-cardinality point is confirmatory; the rest of the ladder is
exploratory.

\textbf{A second temperature.} We repeat the primary elicitation at
temperature 0.7 on the same patches and report the gap to floor at both
settings and the rank correlation between them. With no seed sent, the
second setting is a single stochastic draw whose difference from greedy
mixes the temperature effect with sampling noise, and this design cannot
separate them.

\textbf{Coverage.} Section 3.9.

\textbf{The resolution control on the logit signal.} The first-token
signal takes thousands of distinct values where the elicited object
takes sixteen, so its advantage is open to the same objection the
granularity control answers for the elicited object. The identical
control is run on it in section 4.4. Exploratory.

\textbf{A capacity ladder.} Whether the operative variable is the format
or the model's headroom (Fan, 2026) cannot be settled on one model. We
therefore run both formats on the same 1,800-patch subsets at both
centers on three further vision-capable models served by the same
endpoint under an identical configuration: Gemma-3-27b-it, the
non-medical base model of the same lineage; Gemma-4-31b-it, a later
generation of that lineage; and Mistral-Small-3.1, a different lineage
at comparable scale. Each rung is characterised by its own single-label
accuracy on this task, and the question is whether the nine-class cost
is graded in that. The ladder varies headroom across lineages and
generations rather than within one, and section 6.1 states what that
costs the inference. A fifth vision-capable model was probed and
excluded: it is a reasoning model that emits reasoning tokens before any
visible content, so under the 256-token cap every other run uses it
returns an empty completion on every call, needing 1,073 to 1,765
completion tokens to answer at all. Including it at that cap would
record a truncation artifact as a format effect; including it at a
larger cap would break the identical-decoding control. The ladder
therefore covers non-reasoning models only. Exploratory.

\textbf{A blind-image control.} The mechanistic reading of section 5.2
turns on whether the model is consulting the image, and the standard
control is to remove it. We re-issue the formats on the same subsets
with the patch replaced, holding prompt, model and decoding fixed, under
two treatments. \texttt{blank} sends one uniform mid-grey field to every
call, so the request is byte-identical across the manifest and its parse
rate is one response replicated rather than a proportion over 1,800
draws. \texttt{noise} sends a different seeded random field per patch,
so the image carries no tissue signal but does vary per call, and a rate
measured under it is an estimable proportion. The grey is mid-scale by
choice: the cue block describes \texttt{background} as ``white or
near-empty glass'', so a white blank would test the language prior and
the background prior at once. Exploratory.

\subsection{3.11 Uncertainty, multiplicity, and
cost}\label{uncertainty-multiplicity-and-cost}

Every point estimate carries an interval and the method that produced
it, and every table states the method per row. Three methods appear.
Level AAUCs carry across-seed percentile intervals at S = 200, because
the quantity being resampled is the tie-break. Differences between two
signals on the same patches carry paired patch-level bootstraps at B =
10,000 with a fresh tie-break draw inside each resample, so the interval
carries sampling and tie-break variance together. Proportions carry
Wilson (1927) intervals. Where two signals route different predictions,
the outcome vector travels with its signal and the pairing is at the
patch level.

The minimum detectable effect is read off the paired-bootstrap
distribution of the gap being tested, as (z\_0.975 + z\_0.80) times its
standard deviation. This estimator corrects an earlier one, whose
derivation and consequences are given in Supplementary Material S2.

The confirmatory family was fixed before the data that tests it existed:
six signals against each of two co-primary comparators, twelve tests,
Holm-corrected, with the gap-to-random tests forming a separate family
of the same six. Both raw and adjusted intervals are reported. Analyses
outside that family are labelled exploratory and excluded from the
correction: the accuracy and ranking decomposition, the per-class
breakdown, the cardinality ladder beyond its single matched point, the
capacity ladder, the blind-image control, the decoding sensitivity, the
coverage decomposition, the parse-mode breakdown, the tie-break
variance, and every AUROC here.

Cost is counted in tokens actually spent rather than in queries. We
measured the five-query baseline's denominator directly over 9,000
calls, report query count as a secondary figure, and report how the
ratio degrades under an adaptive stopping rule (Aggarwal et al., 2023).

The resampling unit is the patch. It should be the patient, and the
released PathMNIST arrays carry no slide or patient identifier, so it
cannot be; section 6.5 states what that costs every interval here.

\subsection{3.12 The answer-suppression
control}\label{the-answer-suppression-control}

The variant that separates the prohibition from the cardinality is D4
(section 3.3), and the run that uses it crosses the token budget with
the prompt rather than holding the budget fixed. At a fixed budget the
comparison measures truncation instead: in the 100-patch pilot, 92 of
100 D4 calls ended at the 256-token ceiling, because released from the
prohibition the model reasons about each class in prose before giving
any number and a 256-token completion does not reach the numbers. Three
arms were therefore run on both 1,800-patch subsets, 10,800 calls, with
everything but the named factor at the values every other run in this
paper uses:

{\def\LTcaptype{none} 
\begin{longtable}[]{@{}
  >{\raggedright\arraybackslash}p{(\linewidth - 4\tabcolsep) * \real{0.3333}}
  >{\raggedright\arraybackslash}p{(\linewidth - 4\tabcolsep) * \real{0.3333}}
  >{\raggedright\arraybackslash}p{(\linewidth - 4\tabcolsep) * \real{0.3333}}@{}}
\toprule\noalign{}
\begin{minipage}[b]{\linewidth}\raggedright
Prompt
\end{minipage} & \begin{minipage}[b]{\linewidth}\raggedright
\texttt{max\_tokens}
\end{minipage} & \begin{minipage}[b]{\linewidth}\raggedright
What the arm is for
\end{minipage} \\
\midrule\noalign{}
\endhead
\bottomrule\noalign{}
\endlastfoot
D4 & 256 & the nine-class demand without the prohibition, at the
decoding every other arm uses \\
D4 & 1024 & the same prompt with truncation removed as an explanation \\
D6 & 1024 & the prohibition at the same budget, so the contrast between
the two holds it fixed \\
\end{longtable}
}

The primary contrast is D4 minus D6 at matched budget, both at 1,024
tokens. The secondary is D5 minus D4 at 256, which is the cardinality
effect with suppression absent from both sides. Both are reported on the
scored subset and under the section 3.9 coverage treatment, and an arm
counts as interpretable only if it clears a coverage floor of 0.50 on
the scored basis, fixed before the draw.

\begin{center}\rule{0.5\linewidth}{0.5pt}\end{center}

\section{4. Results}\label{results}

Supplementary Table S1 attributes every number below to the entry that
produced it. Which analyses are confirmatory, and the interval method
each quantity carries, are as stated in section 3.11.

\subsection{4.1 The format changes the
classifier}\label{the-format-changes-the-classifier}

Moving from the three-number request to the nine-class one, under an
identical cue block, costs 14.8 accuracy points on validation and 13.4
at the external center. The nine-class variant's null against its own
random floor is confirmatory; the paired difference against the ranked
top-3 is exploratory and reported outside the Holm correction, because
the preregistered family compares each signal against the five-query
comparators rather than against another elicited variant. Figure 3 shows
what the change does to the accuracy-coverage curve and Figure 4 gives
the difference with intervals.

{\def\LTcaptype{none} 
\begin{longtable}[]{@{}
  >{\raggedright\arraybackslash}p{(\linewidth - 4\tabcolsep) * \real{0.3333}}
  >{\raggedright\arraybackslash}p{(\linewidth - 4\tabcolsep) * \real{0.3333}}
  >{\raggedright\arraybackslash}p{(\linewidth - 4\tabcolsep) * \real{0.3333}}@{}}
\toprule\noalign{}
\begin{minipage}[b]{\linewidth}\raggedright
D6 (nine numbers) minus D5 (ranked top-3), paired patch bootstrap
\end{minipage} & \begin{minipage}[b]{\linewidth}\raggedright
validation, n = 9,163
\end{minipage} & \begin{minipage}[b]{\linewidth}\raggedright
external test, n = 6,599
\end{minipage} \\
\midrule\noalign{}
\endhead
\bottomrule\noalign{}
\endlastfoot
argmax accuracy & \textbf{-0.1475} {[}-0.1594, -0.1353{]} &
\textbf{-0.1341} {[}-0.1484, -0.1196{]} \\
selective-accuracy area (AAUC) & \textbf{-0.1962} {[}-0.2133, -0.1796{]}
& \textbf{-0.2768} {[}-0.2898, -0.2451{]} \\
\end{longtable}
}

Both differences are paired at the patch level on the responses where
both formats parsed, and all four intervals exclude zero. In levels: the
ranked top-3 classifies 0.3817 {[}0.3722, 0.3913{]} of the validation
split correctly and 0.4216 {[}0.4102, 0.4331{]} of the external one,
while the nine-class format manages 0.2223 {[}0.2139, 0.2309{]} and
0.2720 {[}0.2614, 0.2828{]} against a chance level of 0.1111. The
nine-class classifier sits at roughly twice chance, from a baseline more
than half again as high.

The rest of the prompt is identical: same preamble, same nine-line
per-class cue block, same decoding, same token cap, same images. What
differs is the closing block, and it differs in two respects rather than
one (section 3.3). This section reports the difference between the two
prompts as issued; section 4.8 attributes it.

Pairing on the responses both formats parsed removes exactly the patches
the nine-class format fails on, so we measure what that costs. Those
patches concentrate on adipose and background (section 4.2), so the
intersection is not a neutral view of the split. Counting an unparseable
response as an error and comparing both variants on the identical whole
manifest:

{\def\LTcaptype{none} 
\begin{longtable}[]{@{}
  >{\raggedright\arraybackslash}p{(\linewidth - 4\tabcolsep) * \real{0.3333}}
  >{\raggedright\arraybackslash}p{(\linewidth - 4\tabcolsep) * \real{0.3333}}
  >{\raggedright\arraybackslash}p{(\linewidth - 4\tabcolsep) * \real{0.3333}}@{}}
\toprule\noalign{}
\begin{minipage}[b]{\linewidth}\raggedright
Argmax accuracy, unparseable counted as an error, Wilson intervals on
levels
\end{minipage} & \begin{minipage}[b]{\linewidth}\raggedright
validation, n = 10,004
\end{minipage} & \begin{minipage}[b]{\linewidth}\raggedright
external test, n = 7,180
\end{minipage} \\
\midrule\noalign{}
\endhead
\bottomrule\noalign{}
\endlastfoot
ranked top-3 (D5) & 0.3787 {[}0.3693, 0.3883{]} & 0.4201 {[}0.4087,
0.4315{]} \\
nine-class (D6) & 0.2053 {[}0.1975, 0.2133{]} & 0.2510 {[}0.2411,
0.2611{]} \\
difference, paired on the whole manifest & \textbf{-0.1734} {[}-0.1848,
-0.1620{]} & \textbf{-0.1691} {[}-0.1833, -0.1549{]} \\
\end{longtable}
}

The effect is larger under this treatment, not smaller, by 2.6 points on
validation and 3.5 at the external center. The paired figure the paper
leads with is therefore the conservative one.

The cue block matters too, and by a comparable amount. Dropping the cue
block while keeping the ranked top-3 format takes argmax accuracy from
0.3874 {[}0.3651, 0.4103{]} to 0.2083 {[}0.1902, 0.2277{]} on the
validation subset and from 0.3947 {[}0.3723, 0.4176{]} to 0.2267
{[}0.2079, 0.2466{]} on the test subset. Two large prompt-side effects
of similar size, one from what the model is told about the classes and
one from what it is asked to produce; only the second is the subject of
this paper.

The effect on the prediction is sharply nonlinear in what is asked for,
and the effect on routing is not monotone at all. Ordered by how much
the response format demands, the single-query stated confidence produces
a label plus a number, the ranked top-3 produces three, and the
nine-class variant produces nine. Argmax accuracy falls at every step,
0.4081 against 0.3817 on validation and then by 0.1475 more, so almost
the whole of the accuracy cost lands in the last step. Routing value
does something different: absent at one number, present at three, absent
again at nine.

The nine-class score does not clear its own random floor at either
center, which is what makes the collapse a collapse rather than a shift.
In Figure 3 the nine-class curve tracks its own floor and dips below it
across the middle of the coverage range at both centers.

{\def\LTcaptype{none} 
\begin{longtable}[]{@{}
  >{\raggedright\arraybackslash}p{(\linewidth - 4\tabcolsep) * \real{0.3333}}
  >{\raggedright\arraybackslash}p{(\linewidth - 4\tabcolsep) * \real{0.3333}}
  >{\raggedright\arraybackslash}p{(\linewidth - 4\tabcolsep) * \real{0.3333}}@{}}
\toprule\noalign{}
\begin{minipage}[b]{\linewidth}\raggedright
D6 entropy against its own recomputed random floor
\end{minipage} & \begin{minipage}[b]{\linewidth}\raggedright
validation
\end{minipage} & \begin{minipage}[b]{\linewidth}\raggedright
external test
\end{minipage} \\
\midrule\noalign{}
\endhead
\bottomrule\noalign{}
\endlastfoot
gap, scored patches, paired bootstrap & +0.0020 {[}-0.0096, +0.0149{]} &
+0.0095 {[}-0.0043, +0.0277{]} \\
minimum detectable effect & 0.0175 & 0.0228 \\
tie-break variance of the score & 0.0099 & 0.0124 \\
\end{longtable}
}

The validation null is powered to the preregistered resolution: 0.0175
is below the 0.02 practical threshold, so this is a powered null rather
than an absence of evidence. The external-test null is not, at 0.0228.
Both carry a caveat the tie diagnostics on the AAUC rows do not supply:
more than half of what either design could detect is decided inside the
score's tie blocks. The primary variant's tie-break variance on the same
splits is 0.0040 and 0.0042, roughly a fifth of its own resolution, so
this is a property of the nine-class score and not of the protocol.

The collapse is not a model spreading its probability evenly and losing
the argmax to a tie-break. Among the responses the nine-class format
does score, the argmax is attained by more than one class on 0.0017
{[}0.0011, 0.0028{]} of validation and 0.0017 {[}0.0009, 0.0030{]} of
external-center responses, and no scored vector anywhere is near-uniform
in the sense of every stated probability lying within 0.02 of one ninth.
The distributions the format does emit are confident. They are simply
wrong.

Checked at natural prevalence, one number moved. On the class-balanced
subsets the nine-class argmax accuracy is 0.1825 {[}0.1647, 0.2017{]}
and 0.1972 {[}0.1790, 0.2167{]} against 0.2223 and 0.2720 on the whole
splits, with subset minus rest excluding zero on both: the balanced
sample understated the collapse. The claim did not move; the headline
number did.

\subsection{4.2 The format changes what the classifier can
express}\label{the-format-changes-what-the-classifier-can-express}

Per section 3.3, \emph{the nine-class format} means D6 throughout, and
D4 names the same demand without the prohibition. D6 cannot always
produce a distribution, the patches where it cannot are not a random
sample, and the failure is a specific and reproducible response rather
than a formatting accident.

\textbf{What the failure is.} Every nine-class parse failure at both
centers is a sum outside tolerance: none is a missing class, a
duplicated class, a wrong cardinality, or an unrecognised class name. Of
the 764 validation failures, 718 are a vector in which the model names
all nine classes in the requested order and assigns every one of them
0.00, a rate of 0.0718 {[}0.0669, 0.0770{]} over the split and 94.0
percent of the failures; at the external center it is 546 of 554, 98.6
percent, 0.0760 {[}0.0701, 0.0824{]}. D6's characteristic failure is not
a malformed answer but a refusal to place probability anywhere.

The two formats fail for disjoint reasons, which settles whether the
coverage comparison between them is confounded by the asymmetric parse
criterion of section 3.4. Every ranked top-3 failure at both centers, 77
of 10,004 on validation at 0.0077 {[}0.0062, 0.0096{]} and 27 of 7,180
at the external center at 0.0038 {[}0.0026, 0.0055{]}, is the model
naming \texttt{Inflammation}, which is not one of the nine classes. No
top-3 failure is a sum failure and no nine-class failure is a naming
failure, so the sum tolerance the two formats are held to differently
does no work in the gap between them.

\textbf{The equal-footing check.} Renormalising every out-of-tolerance
nine-vector to sum to one moves the nine-class failure rate from 0.0764
{[}0.0713, 0.0817{]} to 0.0718 {[}0.0669, 0.0770{]} on validation and
from 0.0772 {[}0.0712, 0.0836{]} to 0.0760 {[}0.0701, 0.0824{]} at the
external center. The licence reaches 6.0 and 1.4 percent of the failures
and cannot reach a vector of nine zeros. The concentration survives it:
adipose stays at 0.4582 {[}0.4282, 0.4886{]} and 0.2915 {[}0.2678,
0.3164{]}, background at 0.1580 {[}0.1372, 0.1812{]} and 0.1511
{[}0.1286, 0.1768{]}.

Failure concentrates on the two visually simplest classes. Wilson
intervals on proportions:

{\def\LTcaptype{none} 
\begin{longtable}[]{@{}
  >{\raggedright\arraybackslash}p{(\linewidth - 4\tabcolsep) * \real{0.3333}}
  >{\raggedright\arraybackslash}p{(\linewidth - 4\tabcolsep) * \real{0.3333}}
  >{\raggedright\arraybackslash}p{(\linewidth - 4\tabcolsep) * \real{0.3333}}@{}}
\toprule\noalign{}
\begin{minipage}[b]{\linewidth}\raggedright
D6 parse-failure rate
\end{minipage} & \begin{minipage}[b]{\linewidth}\raggedright
validation
\end{minipage} & \begin{minipage}[b]{\linewidth}\raggedright
external test
\end{minipage} \\
\midrule\noalign{}
\endhead
\bottomrule\noalign{}
\endlastfoot
adipose & \textbf{0.4592} {[}0.4291, 0.4895{]} & \textbf{0.2922}
{[}0.2685, 0.3172{]} \\
background & \textbf{0.1675} {[}0.1462, 0.1912{]} & \textbf{0.1547}
{[}0.1319, 0.1806{]} \\
\end{longtable}
}

All eighteen per-class intervals exclude their split's overall rate, and
unlike the accuracy figure in section 4.1 this pattern is not a sampling
artefact: subset minus the rest of the same split contains zero for both
classes on both splits.

Counting those failures changes the verdict on the nine-class score at
both centers. Every null in section 4.1 is computed on the 92.4 and 92.3
percent of each split the format could score, which is depleted of
exactly the two classes it fails on. Under the coverage treatment of
section 3.9, with difference rows computed as paired resamples on a
common index rather than as differences of the two levels:

{\def\LTcaptype{none} 
\begin{longtable}[]{@{}
  >{\raggedright\arraybackslash}p{(\linewidth - 8\tabcolsep) * \real{0.2000}}
  >{\raggedright\arraybackslash}p{(\linewidth - 8\tabcolsep) * \real{0.2000}}
  >{\raggedright\arraybackslash}p{(\linewidth - 8\tabcolsep) * \real{0.2000}}
  >{\raggedright\arraybackslash}p{(\linewidth - 8\tabcolsep) * \real{0.2000}}
  >{\raggedright\arraybackslash}p{(\linewidth - 8\tabcolsep) * \real{0.2000}}@{}}
\toprule\noalign{}
\begin{minipage}[b]{\linewidth}\raggedright
D6 gap to floor
\end{minipage} & \begin{minipage}[b]{\linewidth}\raggedright
scored subset
\end{minipage} & \begin{minipage}[b]{\linewidth}\raggedright
pessimistic
\end{minipage} & \begin{minipage}[b]{\linewidth}\raggedright
difference
\end{minipage} & \begin{minipage}[b]{\linewidth}\raggedright
verdict
\end{minipage} \\
\midrule\noalign{}
\endhead
\bottomrule\noalign{}
\endlastfoot
validation & +0.0020 {[}-0.0096, +0.0149{]} & \textbf{+0.0154}
{[}+0.0077, +0.0301{]} & +0.0158 {[}+0.0029, +0.0304{]} & changes \\
external test & +0.0095 {[}-0.0043, +0.0277{]} & \textbf{+0.0337}
{[}+0.0159, +0.0461{]} & +0.0191 {[}+0.0021, +0.0367{]} & changes \\
\end{longtable}
}

The primary variant does not move under the identical treatment: +0.0025
{[}-0.0119, +0.0168{]} and +0.0011 {[}-0.0153, +0.0178{]}. With 27 to 77
failures there is nothing to earn. Neither would the class-balanced
subsets have found this, since both treatments contain zero there at
both centers.

The decomposition is the finding (Figure 5). Appending known-incorrect
patches at the bottom of a ranking raises selective-accuracy area above
the floor even when the score above them is pure noise, so the
pessimistic gap splits exactly into a deferral term and a residual
(section 3.9). These two terms are computed on one patch set and
therefore do sum exactly:

{\def\LTcaptype{none} 
\begin{longtable}[]{@{}
  >{\raggedright\arraybackslash}p{(\linewidth - 4\tabcolsep) * \real{0.3333}}
  >{\raggedright\arraybackslash}p{(\linewidth - 4\tabcolsep) * \real{0.3333}}
  >{\raggedright\arraybackslash}p{(\linewidth - 4\tabcolsep) * \real{0.3333}}@{}}
\toprule\noalign{}
\begin{minipage}[b]{\linewidth}\raggedright
D6, full scale, paired bootstrap on one patch set
\end{minipage} & \begin{minipage}[b]{\linewidth}\raggedright
the deferral rule alone
\end{minipage} & \begin{minipage}[b]{\linewidth}\raggedright
what the entropy adds on top of it
\end{minipage} \\
\midrule\noalign{}
\endhead
\bottomrule\noalign{}
\endlastfoot
validation & \textbf{+0.0151} {[}+0.0052, +0.0272{]}, excludes zero &
\textbf{+0.0003} {[}-0.0085, +0.0136{]}, contains zero, at a resolution
of 0.0160 \\
external test & \textbf{+0.0205} {[}+0.0058, +0.0341{]}, excludes zero &
\textbf{+0.0132} {[}-0.0040, +0.0256{]}, contains zero, at a resolution
of 0.0213 \\
\end{longtable}
}

The two centers say different things about the residual and the
difference is the design's, not the model's. On validation the residual
is three ten-thousandths of an AAUC point against a design that could
have resolved 0.0160, below the 0.02 practical threshold: a powered
null. At the external center the residual is larger, its interval still
contains zero, and the design could not resolve better than 0.0213,
which is above the threshold. That center's result is not
distinguishable, and the validation result is the only equivalence-grade
null in this study.

\subsection{4.3 The format changes routing against the baseline it
replaces}\label{the-format-changes-routing-against-the-baseline-it-replaces}

Here the cue block is held fixed and only the response format changes,
and this is the only comparator comparison that exists at natural
prevalence on a whole split.

{\def\LTcaptype{none} 
\begin{longtable}[]{@{}
  >{\raggedright\arraybackslash}p{(\linewidth - 4\tabcolsep) * \real{0.3333}}
  >{\raggedright\arraybackslash}p{(\linewidth - 4\tabcolsep) * \real{0.3333}}
  >{\raggedright\arraybackslash}p{(\linewidth - 4\tabcolsep) * \real{0.3333}}@{}}
\toprule\noalign{}
\begin{minipage}[b]{\linewidth}\raggedright
Elicited top-3 entropy minus the single-query stated confidence, AAUC,
paired bootstrap
\end{minipage} & \begin{minipage}[b]{\linewidth}\raggedright
validation, n = 9,927
\end{minipage} & \begin{minipage}[b]{\linewidth}\raggedright
external test, n = 7,153
\end{minipage} \\
\midrule\noalign{}
\endhead
\bottomrule\noalign{}
\endlastfoot
stated-label extraction (primary) & \textbf{+0.0244} {[}+0.0118,
+0.0476{]} & \textbf{+0.1668} {[}+0.1485, +0.1889{]} \\
sweep-based extraction & +0.0308 {[}+0.0180, +0.0539{]} & +0.1797
{[}+0.1619, +0.2025{]} \\
\end{longtable}
}

Both gaps exclude zero under either convention, so this verdict does not
depend on section 3.7. In levels, the elicited object reaches 0.4329
{[}0.4312, 0.4344{]} and 0.5618 {[}0.5598, 0.5640{]}; the scalar it
replaces reaches 0.4014 {[}0.3937, 0.4083{]} and 0.3920 {[}0.3817,
0.4006{]} under the same extraction.

The elicited object's full-scale validation AAUC of 0.4329, at n = 9,927
with a paired bootstrap, is equal to four decimals to the five-query
agreement comparator's reference value of 0.4329, at n = 1,800 with an
across-seed percentile interval. These are different signals on
different patch sets under different interval methods, and the agreement
is arithmetic coincidence rather than transcription.

The scalar the elicited object replaces has nothing to give. Its AUROC
against per-patch correctness on the full validation split is 0.4975
{[}0.4938, 0.5012{]}, which is chance, and its gap to its own recomputed
random floor on validation is -0.0049, contained in the level row. This
is a regime and not a general claim about verbalized confidence: at the
external center the same scalar does carry a small positive gap of
+0.0397, contained in its own level row.

Both preregistered gates pass and neither moves under either extraction
convention. The first is the elicited object's gap to its recomputed
random floor on the validation 1,800-patch subset, +0.0880 {[}+0.0514,
+0.1142{]}; the second is the same quantity on the test subset, +0.1190
{[}+0.0942, +0.1580{]}. At full scale the same quantity is +0.0493
{[}+0.0377, +0.0649{]} and +0.1337 {[}+0.1247, +0.1559{]}.

One quantity here runs the other way. The elicited object's
\emph{accuracy} advantage over the scalar does not survive the move to
natural prevalence. On validation it is -0.0264 under stated-label
extraction and -0.0224 under sweep-based, against subset values around
+0.049; at the external center it is +0.0685 and +0.0787. A sign change
between the class-balanced sample and the whole split, driven by
re-weighting classes on which the model's competence is extremely
uneven. We proposed this quantity as a secondary headline and it did not
survive; section 3.5 states the scoping rule that followed.

\subsection{4.4 Routing in context: what the elicited object loses
to}\label{routing-in-context-what-the-elicited-object-loses-to}

\subsubsection{4.4.1 The five-query comparator, a closed
non-finding}\label{the-five-query-comparator-a-closed-non-finding}

Against the two preregistered five-query comparators the elicited object
is not distinguishable at either center, at a realised minimum
detectable effect of 0.0517 on validation and 0.0526 at the external
center, both on the pooled subset-plus-extension patch sets under
stated-label extraction, against the 0.02 practical threshold.

{\def\LTcaptype{none} 
\begin{longtable}[]{@{}
  >{\raggedright\arraybackslash}p{(\linewidth - 4\tabcolsep) * \real{0.3333}}
  >{\raggedright\arraybackslash}p{(\linewidth - 4\tabcolsep) * \real{0.3333}}
  >{\raggedright\arraybackslash}p{(\linewidth - 4\tabcolsep) * \real{0.3333}}@{}}
\toprule\noalign{}
\begin{minipage}[b]{\linewidth}\raggedright
Elicited top-3 entropy minus the five-query agreement score, paired
bootstrap
\end{minipage} & \begin{minipage}[b]{\linewidth}\raggedright
sweep-based extraction
\end{minipage} & \begin{minipage}[b]{\linewidth}\raggedright
stated-label extraction
\end{minipage} \\
\midrule\noalign{}
\endhead
\bottomrule\noalign{}
\endlastfoot
validation, n = 1,781 & +0.0326 {[}-0.0054, +0.0720{]} & -0.0306
{[}-0.0648, +0.0111{]} \\
external test, n = 1,786 & +0.0681 {[}+0.0259, +0.1061{]}, Holm 0.00540,
rejects & +0.0118 {[}-0.0265, +0.0530{]}, Holm 1.00000 \\
validation, pooled n = 1,981 & +0.0301 {[}-0.0117, +0.0619{]} & -0.0328
{[}-0.0699, +0.0013{]} \\
external test, pooled n = 2,124 & +0.0615 {[}+0.0245, +0.0999{]}, Holm
0.0156, rejects & +0.0043 {[}-0.0282, +0.0457{]} \\
\end{longtable}
}

The external-center superiority exists only under sweep-based
extraction: under a comparator whose labels are extracted the same way
as the elicited object's, it does not survive at either sample size, and
the cause is the extraction convention rather than endpoint drift
(Supplementary Material S2).

This is a permanent non-finding rather than a deferral. The resolution
the two whole splits could deliver is 0.0230 and 0.0286, and reaching
0.02 would need more patches than either split contains (section 3.8),
so there is no budget at which this comparison becomes decidable on this
dataset.

\subsubsection{4.4.2 The first-token signal outperforms every elicited
object}\label{the-first-token-signal-outperforms-every-elicited-object}

Figure 6 puts every routing comparison on one axis; its top two rows are
each a signal against its own floor rather than a head-to-head. The
normalized entropy of the top-5 logits at the first content-bearing
answer token arrives in the same response, at zero marginal token cost,
wherever the endpoint returns logprobs.

{\def\LTcaptype{none} 
\begin{longtable}[]{@{}
  >{\raggedright\arraybackslash}p{(\linewidth - 4\tabcolsep) * \real{0.3333}}
  >{\raggedright\arraybackslash}p{(\linewidth - 4\tabcolsep) * \real{0.3333}}
  >{\raggedright\arraybackslash}p{(\linewidth - 4\tabcolsep) * \real{0.3333}}@{}}
\toprule\noalign{}
\begin{minipage}[b]{\linewidth}\raggedright
Full scale, gap to own recomputed random floor, paired bootstrap
\end{minipage} & \begin{minipage}[b]{\linewidth}\raggedright
validation
\end{minipage} & \begin{minipage}[b]{\linewidth}\raggedright
external test
\end{minipage} \\
\midrule\noalign{}
\endhead
\bottomrule\noalign{}
\endlastfoot
first-token entropy & \textbf{+0.1820} {[}+0.1722, +0.1974{]} &
\textbf{+0.1775} {[}+0.1700, +0.1984{]} \\
elicited top-3 entropy & +0.0493 {[}+0.0377, +0.0649{]} & +0.1337
{[}+0.1247, +0.1559{]} \\
first-token ranking efficiency & 0.5026 {[}0.4802, 0.5246{]} & 0.5059
{[}0.4819, 0.5296{]} \\
elicited ranking efficiency & 0.1394 {[}0.1133, 0.1653{]} & 0.3851
{[}0.3556, 0.4143{]} \\
\end{longtable}
}

The margin is much narrower at the external center than at validation,
and the discrimination numbers show it plainly: AUROC 0.7089 {[}0.6967,
0.7211{]} against 0.6767 {[}0.6646, 0.6886{]} at the external center,
where on validation it is 0.7172 {[}0.7066, 0.7280{]} against 0.5943
{[}0.5833, 0.6054{]}.

The logit signal also beats the five-query baseline the elicited object
could not, on one call against five, at both centers and under all three
versions of the comparator. These rows are at n = 1,787 rather than the
n = 1,786 of section 4.4.1 because the first-token signal is available
on one external-center patch whose elicited response failed to parse.

{\def\LTcaptype{none} 
\begin{longtable}[]{@{}
  >{\raggedright\arraybackslash}p{(\linewidth - 4\tabcolsep) * \real{0.3333}}
  >{\raggedright\arraybackslash}p{(\linewidth - 4\tabcolsep) * \real{0.3333}}
  >{\raggedright\arraybackslash}p{(\linewidth - 4\tabcolsep) * \real{0.3333}}@{}}
\toprule\noalign{}
\begin{minipage}[b]{\linewidth}\raggedright
First-token entropy minus the five-query agreement score, paired
bootstrap
\end{minipage} & \begin{minipage}[b]{\linewidth}\raggedright
validation, n = 1,781
\end{minipage} & \begin{minipage}[b]{\linewidth}\raggedright
external test, n = 1,787
\end{minipage} \\
\midrule\noalign{}
\endhead
\bottomrule\noalign{}
\endlastfoot
original traces, sweep-based extraction & +0.1259 {[}+0.0965, +0.1735{]}
& +0.1135 {[}+0.0776, +0.1575{]} \\
re-issued calls, sweep-based extraction & +0.1269 {[}+0.0982, +0.1749{]}
& +0.1138 {[}+0.0777, +0.1579{]} \\
re-issued calls, stated-label extraction & +0.0662 {[}+0.0402,
+0.1091{]} & +0.0594 {[}+0.0273, +0.1002{]} \\
\end{longtable}
}

Against the second comparator, predictive entropy over the five labels,
the stated-label column is +0.0662 {[}+0.0401, +0.1087{]} and +0.0585
{[}+0.0268, +0.0996{]}. Twelve intervals of twelve exclude zero, and
correcting the comparator roughly halves the free signal's margin
without removing it.

The advantage is not resolution. The first-token score takes 9,223
distinct values on validation where the elicited object takes 16, so its
advantage could have been resolution rather than information. Quantized
to 16 levels by equal-frequency binning, matched to the elicited
object's own cardinality, it loses nothing: +0.0012 {[}-0.0019,
+0.0038{]} on validation at a resolution of 0.0041, and +0.0002
{[}-0.0039, +0.0005{]} at the external center at 0.0032. Both nulls are
powered an order of magnitude finer than the practical threshold.
Degradation only becomes resolvable at five bins and below, and at two
bins the signal still reaches 0.4821 {[}0.4752, 0.4891{]} on validation,
above the elicited object's full-resolution 0.4329 {[}0.4312, 0.4344{]}
on the same split. That asymmetry does not hold at the external center.

What the elicited object is, then, is a different signal rather than a
coarse readout of the logits. On MedGemma the two correlate at Spearman
-0.1818 {[}-0.2004, -0.1625{]} and -0.3089 {[}-0.3293, -0.2889{]}, the
negative sign following from opposed conventions, so the comparable
quantity is the magnitude. That is a MedGemma statement: on
Gemma-3-27b-it the two correlate at -0.6243 {[}-0.6560, -0.5908{]}, a
much tighter coupling.

\subsubsection{4.4.3 Cost, measured}\label{cost-measured}

The five-query baseline costs 5.84 times the input tokens, 5.24 times
the output tokens and 5.79 times the total tokens of one elicited query,
measured from the endpoint's own usage field over five sealed runs and
9,000 calls, against 5.00 times the queries. The token accounting
slightly favours the single query rather than eroding it. Under an
adaptive stopping rule at an effective K of 2 the total-token ratio
falls to 2.32 times and at K = 1.8 to 2.09 times, so the saving is
between two and six times depending on the stopping rule, not five
times.

\subsection{4.5 Mechanism}\label{mechanism}

The elicited object's advantage over the five-query comparator is not
resolution. Quantized by equal-frequency binning to five levels, the
cardinality of the five-query agreement score itself, elicited entropy
still beats that comparator at the external center by +0.0539
{[}+0.0050, +0.0871{]} and beats predictive entropy by +0.0534
{[}+0.0043, +0.0866{]}, both excluding zero. This is the confirmatory
matched-cardinality control, and its scope is narrow: on validation the
same rung contains zero, as the unquantized comparison does, so the
control shows that the \emph{external-center} advantage is not a
resolution artefact rather than that there is an advantage at both
centers.

The elicited entropy is not a restatement of the scalar it replaces. At
full scale the two correlate at Spearman -0.0249 {[}-0.0411, -0.0088{]}
on validation and +0.0164 {[}+0.0010, +0.0314{]} at the external center.
At n near 10,000 an interval on a correlation of 0.02 no longer contains
zero, so the reportable statement is that the association is negligible
in magnitude, not that it is indistinguishable from zero.

Entropy was chosen over max probability because it never reads the
maximum, and on a three-element vector plus a residual the two are
nearly rank-identical, so the choice makes little difference to any
result here: at full scale they correlate at +0.9833 {[}+0.9823,
+0.9842{]} on validation and +0.9762 {[}+0.9747, +0.9776{]} at the
external center.

The cross-center instability is real, survives every manipulation aimed
at it, and we cannot explain it. The elicited object's ranking
efficiency is 0.1394 {[}0.1133, 0.1653{]} on validation and 0.3851
{[}0.3556, 0.4143{]} at the external center, two intervals that do not
overlap. It is not resolution: test minus validation ranking efficiency
is +0.2457 {[}+0.2051, +0.2854{]} unquantized and stays at +0.1637
{[}+0.1241, +0.2041{]} even at two bins and +0.2408 {[}+0.2023,
+0.2771{]} at sixteen. And it is not a property of the model on those
patches, because the first-token signal read from the same calls is
stable across the same two centers, 0.5026 {[}0.4802, 0.5246{]} against
0.5059 {[}0.4819, 0.5296{]}. Whatever is unstable is in the elicited
signal.

One rung of the cardinality ladder beats the unquantized score at both
centers. Quantizing elicited entropy to sixteen requested bins, which
realises seven actual levels because equal-frequency binning collapses
on a score with large tie blocks, gives 0.4478 {[}0.4439, 0.4511{]}
against the unquantized 0.4329 {[}0.4312, 0.4344{]} on validation and
0.5748 {[}0.5715, 0.5780{]} against 0.5618 {[}0.5598, 0.5640{]} at the
external center. The extra resolution in the raw score is costing
ranking quality. Exploratory.

\subsection{4.6 The capacity ladder: how much of the cost is the
model}\label{the-capacity-ladder-how-much-of-the-cost-is-the-model}

Under the capacity account of Fan (2026) a structural demand is free for
a model with capacity to spare and expensive for one near its limits, so
the cost should be graded in how well the model does the task. Four
vision-capable models were run on the same 1,800 patches at both
centers, under the same prompts byte for byte, the same decoding and the
same parser, with each rung placed on the ladder by its own accuracy
under the ranked top-3.

{\def\LTcaptype{none} 
\begin{longtable}[]{@{}
  >{\raggedright\arraybackslash}p{(\linewidth - 6\tabcolsep) * \real{0.2500}}
  >{\raggedright\arraybackslash}p{(\linewidth - 6\tabcolsep) * \real{0.2500}}
  >{\raggedright\arraybackslash}p{(\linewidth - 6\tabcolsep) * \real{0.2500}}
  >{\raggedright\arraybackslash}p{(\linewidth - 6\tabcolsep) * \real{0.2500}}@{}}
\toprule\noalign{}
\begin{minipage}[b]{\linewidth}\raggedright
Model, ordered by its own ranked top-3 accuracy
\end{minipage} & \begin{minipage}[b]{\linewidth}\raggedright
D5
\end{minipage} & \begin{minipage}[b]{\linewidth}\raggedright
D6
\end{minipage} & \begin{minipage}[b]{\linewidth}\raggedright
D6 minus D5, paired on the whole manifest
\end{minipage} \\
\midrule\noalign{}
\endhead
\bottomrule\noalign{}
\endlastfoot
\textbf{Validation} & & & \\
Gemma-3-27b-it & 0.3417 & 0.3000 & -0.0417 {[}-0.0639, -0.0194{]} \\
MedGemma-27b-it & 0.3833 & 0.1689 & \textbf{-0.2144} {[}-0.2389,
-0.1894{]} \\
Mistral-Small-3.1 & 0.3944 & 0.3556 & -0.0389 {[}-0.0533, -0.0239{]} \\
Gemma-4-31b-it & 0.5728 & 0.5450 & -0.0278 {[}-0.0428, -0.0133{]} \\
\textbf{External test} & & & \\
Gemma-3-27b-it & 0.3372 & 0.2844 & -0.0528 {[}-0.0756, -0.0300{]} \\
Mistral-Small-3.1 & 0.3878 & 0.3600 & -0.0278 {[}-0.0422, -0.0133{]} \\
MedGemma-27b-it & 0.3917 & 0.1867 & \textbf{-0.2050} {[}-0.2311,
-0.1794{]} \\
Gemma-4-31b-it & 0.4967 & 0.4739 & -0.0228 {[}-0.0361, -0.0106{]} \\
\end{longtable}
}

Accuracies are on the whole manifest with an unparseable response
counted as an error, which is the only basis on which models with
different coverage are comparable; differences are paired resamples on
the shared patch index. All eight differences exclude zero, so every
model pays something for the nine-class demand.

The cost is not graded in accuracy. Across a range of ranked top-3
accuracy from 0.3417 to 0.5728 on validation, three of the four models
sit in a narrow band between -0.0278 and -0.0417, and the fourth is five
times worse. The pattern repeats at the external center. A capacity
account predicts a monotone relationship between headroom and cost; what
the ladder shows is a flat band and one outlier.

Mistral-Small-3.1 against MedGemma-27b-it makes this sharp, because the
two are matched on the variable the capacity account turns on. Their
ranked top-3 accuracies bracket each other at both centers, 0.3944
against 0.3833 on validation and 0.3878 against 0.3917 at the external
center, so on the accuracy proxy for headroom they are the same model
twice. Their nine-class costs differ by a factor of 5.5 on validation
and 7.4 at the external center. Two models with the same amount to lose
do not lose the same amount.

The coverage numbers separate the same model even more sharply. The
nine-class format's parse rate is 0.9978 and 0.9994 for Gemma-3, 0.9933
and 0.9922 for Mistral-Small, and 0.9906 and 0.9833 for Gemma-4, against
0.9256 and 0.9467 for MedGemma. The nine-zeros response of section 4.2
is close to absent outside the one model: on validation MedGemma emits
it on 134 of 1,800 subset patches where the next highest is 13.

This establishes that the cost is not a smooth function of a model's
competence at the task, so the format effect is not headroom under
another name. It does not establish what the cost \emph{is} a function
of: the one model that pays is also the only one of the four with
medical post-training, which a design with one medical model cannot
test, and the four differ in lineage, generation and training data as
well as in accuracy. Section 6.1 states what a ladder built this way can
support.

\subsection{4.7 The blind-image control: what the format does with
nothing to
see}\label{the-blind-image-control-what-the-format-does-with-nothing-to-see}

Each variant was re-issued on the same 1,800-patch subsets with the
patch replaced, everything else held fixed, under two treatments
(section 3.10). D4 is included because D5 against D6 alone tests the
prohibition and the cardinality together.

{\def\LTcaptype{none} 
\begin{longtable}[]{@{}
  >{\raggedright\arraybackslash}p{(\linewidth - 6\tabcolsep) * \real{0.2500}}
  >{\raggedright\arraybackslash}p{(\linewidth - 6\tabcolsep) * \real{0.2500}}
  >{\raggedright\arraybackslash}p{(\linewidth - 6\tabcolsep) * \real{0.2500}}
  >{\raggedright\arraybackslash}p{(\linewidth - 6\tabcolsep) * \real{0.2500}}@{}}
\toprule\noalign{}
\begin{minipage}[b]{\linewidth}\raggedright
Image replaced, parse rate and all-zeros rate, Wilson intervals
\end{minipage} & \begin{minipage}[b]{\linewidth}\raggedright
ranked top-3 (D5)
\end{minipage} & \begin{minipage}[b]{\linewidth}\raggedright
nine-class, prohibition (D6)
\end{minipage} & \begin{minipage}[b]{\linewidth}\raggedright
nine-class, no prohibition (D4)
\end{minipage} \\
\midrule\noalign{}
\endhead
\bottomrule\noalign{}
\endlastfoot
uniform mid-grey, validation, parse rate & 1.0000 {[}0.9979, 1.0000{]} &
\textbf{0.0000} {[}0.0000, 0.0021{]} & 0.8022 {[}0.7832, 0.8200{]} \\
uniform mid-grey, validation, all-zeros & n/a & \textbf{1.0000}
{[}0.9979, 1.0000{]} & \textbf{0.0000} {[}0.0000, 0.0021{]} \\
uniform mid-grey, validation, accuracy & 0.1111 & 0.0000 & n/a \\
seeded noise, validation, parse rate & 1.0000 {[}0.9979, 1.0000{]} &
0.9994 {[}0.9969, 0.9999{]} & 0.6150 {[}0.5923, 0.6372{]} \\
seeded noise, validation, all-zeros & n/a & 0.0006 {[}0.0001, 0.0031{]}
& 0.0000 {[}0.0000, 0.0021{]} \\
seeded noise, validation, accuracy & 0.1111 & 0.1111 & n/a \\
seeded noise, external test, parse rate & 1.0000 {[}0.9979, 1.0000{]} &
0.9994 {[}0.9969, 0.9999{]} & 0.6117 {[}0.5889, 0.6339{]} \\
seeded noise, external test, all-zeros & n/a & 0.0006 {[}0.0001,
0.0031{]} & 0.0000 {[}0.0000, 0.0021{]} \\
seeded noise, external test, accuracy & 0.1111 & 0.1111 & n/a \\
\end{longtable}
}

An all-zeros rate is defined only where nine numbers were requested, so
it is not reported for the ranked top-3.

On a uniform field D6 emits the nine-zeros vector on every call and the
ranked top-3 format emits a well-formed answer. The response the
nine-class format returns is byte-identical to the one it returns on the
real patches it fails, and it is the only response the run produced: one
distinct string across 1,800 calls. The ranked top-3 format, given the
same blank field, returns
\texttt{background\ 0.70,\ adipose\ 0.20,\ debris\ 0.10} on 1,781 of its
1,800 calls, which is a sensible prior over an empty slide. Same model,
same cue block, same absence of evidence: one format falls back on a
prior and the other produces nothing.

The trigger is a uniform field, not the absence of information. Under
seeded noise, which carries no tissue signal but plenty of visual
variance, D6 parses on 0.9994 of calls at both centers and returns a
confident distribution. The failure is not the model declining to answer
when the image is uninformative; it is the model declining to answer
when the image is featureless. That is the property adipose and
background share, and it is why the failures concentrate there.

It also rules out a language-prior account, under which the nine-class
demand makes the model generate a plausible nine-vector without
consulting the image: given no image to consult, D6 produces zeros
rather than a plausible vector, and the format that falls back on a
prior is the top-3 one.

The refusal belongs to the prohibition rather than to the nine-class
demand. D4 makes the identical nine-class request without the closing
sentence. Given the same uniform mid-grey field it emits the zero vector
on no call at all, 0.0000 {[}0.0000, 0.0021{]}, against D6's 1.0000 on
the identical input, and it parses on 0.8022 {[}0.7832, 0.8200{]} of
those calls, close to the 0.8622 it manages on real patches. So a
featureless image does not by itself produce the refusal, and the
prohibition does not by itself produce it either, since D6 answers most
ordinary patches. It takes both, and the sentence that carries the
accuracy cost in section 4.8 is the same sentence that, on a featureless
field, moves the model from answering badly to not answering at all.

The noise arms are the control rather than the test, and neither variant
refuses under seeded noise. What they do show is that D4's format
compliance falls to 0.6150 {[}0.5923, 0.6372{]} and 0.6117 {[}0.5889,
0.6339{]} on a textured but meaningless field, against 0.9994 for D6,
and that its 1,581 and 1,575 distinct responses across 1,800 calls make
it by far the least reproducible of the three. A variant that is allowed
to reason at length pays for it in stability.

Two limits on how far this can be pushed. The uniform-field run sends a
byte-identical request on every call, so its rate of 1.0000 is one
response replicated rather than a proportion over 1,800 independent
draws. And an accuracy of 0.1111 on a class-stratified subset is what a
constant prediction scores, so the accuracy rows record that neither
format retains any discrimination without the image rather than showing
that both retain the same amount.

\subsection{4.8 The answer-suppression
control}\label{the-answer-suppression-control-1}

D4 removes the prohibition and holds the requested cardinality (section
3.12). The result replicates at both centers, and it changes what the
difference of section 4.1 is a difference between.

\subsubsection{4.8.1 The token budget does nothing for D6 and everything
for
D4}\label{the-token-budget-does-nothing-for-d6-and-everything-for-d4}

Given four times the room, D6's format compliance is 0.9256 {[}0.9125,
0.9368{]} against 0.9256 at the smaller cap, its argmax accuracy is
0.1783 against 0.1825, and it emits 86.1 completion tokens per call at
both caps with a longest response of 87. At the external center the same
holds: 0.9461 {[}0.9347, 0.9556{]} and 0.1955. Offered the space, the
nine-class prompt with the prohibition does not use one token of it. It
is not running out of room; it has been told not to use any.

D4 behaves in the opposite way. At the 256-token cap every other run in
this paper uses, its compliance is 0.0994 {[}0.0865, 0.1141{]} and
0.1289 {[}0.1142, 0.1452{]}, with 1,626 and 1,573 of 1,800 responses
ending at the ceiling. At 1,024 the truncation is gone, 12 and 9
responses, and compliance rises to 0.8622 {[}0.8455, 0.8774{]} and
0.8628 {[}0.8461, 0.8779{]}, clearing at both centers the 0.50 floor
fixed before the draw. D4 spends 480.4 and 470.0 completion tokens per
call against D5's 36.6, and section 5.4 prices that.

\subsubsection{4.8.2 The prohibition against the
cardinality}\label{the-prohibition-against-the-cardinality}

Three contrasts separate the two factors, each computed on the patches
both of its arms parsed:

{\def\LTcaptype{none} 
\begin{longtable}[]{@{}
  >{\raggedright\arraybackslash}p{(\linewidth - 4\tabcolsep) * \real{0.3333}}
  >{\raggedright\arraybackslash}p{(\linewidth - 4\tabcolsep) * \real{0.3333}}
  >{\raggedright\arraybackslash}p{(\linewidth - 4\tabcolsep) * \real{0.3333}}@{}}
\toprule\noalign{}
\begin{minipage}[b]{\linewidth}\raggedright
Contrast in argmax accuracy, responses both arms parsed
\end{minipage} & \begin{minipage}[b]{\linewidth}\raggedright
validation 1,800
\end{minipage} & \begin{minipage}[b]{\linewidth}\raggedright
external test 1,800
\end{minipage} \\
\midrule\noalign{}
\endhead
\bottomrule\noalign{}
\endlastfoot
\textbf{D4 minus D6}, matched cardinality and budget: the prohibition &
\textbf{+0.1878} {[}+0.1623, +0.2134{]} & \textbf{+0.1911} {[}+0.1648,
+0.2181{]} \\
\textbf{D5 minus D4}, prohibition absent from both: the cardinality &
-0.0013 {[}-0.0300, +0.0280{]} & -0.0214 {[}-0.0494, +0.0071{]} \\
D5 minus D6, the pair section 4.1 reports & +0.1955 {[}+0.1700,
+0.2216{]} & +0.1846 {[}+0.1586, +0.2118{]} \\
\end{longtable}
}

Two properties of the table qualify it. First, the cardinality row
compares D5 at a 256-token cap against D4 at 1,024, and that asymmetry
does not smuggle the budget into the contrast: no D5 response anywhere
in this study comes near its cap, the longest being 43 tokens on
validation and 42 at the external center, so raising D5's budget could
change nothing. Second, the three rows sit on three different patch
intersections and therefore do not sum. They are computed on 1,533,
1,448 and 1,647 patches on validation and 1,540, 1,481 and 1,690 at the
external center, because each pair of arms fails on a different set.
Adding the first two gives +0.1865 against a directly measured +0.1955
on validation, and +0.1697 against +0.1846 externally.

Removing one sentence and changing nothing else recovers essentially the
whole of the effect at both centers, and what is left for the nine-class
request itself to explain contains zero at both. D4 reaches 0.3821 and
0.3986 against D5's 0.3874 and 0.3947 and D6's 0.1783 and 0.1955.

\textbf{A declared deviation.} The preregistration named the secondary
contrast as D5 minus D4 \emph{at the 256-token cap}. At that cap D4
parses one response in ten, so on the scored basis that contrast is
computed on 179 and 232 patches selected for being the ones the model
happened to answer briefly, and it is not interpretable as a cardinality
effect: it reads -0.0782 {[}-0.1676, +0.0168{]} and +0.0259 {[}-0.0431,
+0.0948{]} at minimum detectable effects of 0.0922 and 0.0711. The row
in the table above is the same comparison against D4 at the larger cap,
where coverage clears the preregistered floor of section 3.12. Which arm
that floor would exclude was not anticipated.

Three qualifications bound the result. The cardinality row is a tight
null and not a powered one: its realised minimum detectable effects are
0.0297 and 0.0279, above the 0.02 threshold, so a cost of up to about
three points on validation and five at the external center cannot be
excluded, and on the external split the point estimate runs slightly in
D4's favour. It is a null on the scored basis only: under the section
3.9 coverage treatment D5 does beat D4, by +0.0539 {[}+0.0267,
+0.0811{]} and +0.0478 {[}+0.0211, +0.0744{]}, and that gap is coverage
rather than accuracy, since D4 complies at 0.86 against D5's 0.99. And
the primary contrast survives the same treatment, +0.1644 {[}+0.1400,
+0.1883{]} and +0.1589 {[}+0.1344, +0.1828{]}. The nine-class request
costs coverage; the prohibition costs the accuracy.

\subsubsection{4.8.3 A degenerate classifier rather than a degraded
one}\label{a-degenerate-classifier-rather-than-a-degraded-one}

Accuracy alone cannot tell a model that is spread thin from one that has
stopped classifying, so we report the share of scored responses landing
on the single most-predicted class. D6 puts 0.8115 of its validation
predictions and 0.7576 of its external-center predictions on one class
and never uses two of the nine at all; at the larger budget it is 0.8151
and 0.7663, unmoved. D4 sits at 0.2107 and 0.2589 and uses all nine, and
D5 at 0.2628 and 0.2811. What the prohibition produces is not a worse
classifier but a model that has stopped choosing, which is what the
instruction literally asks for.

\subsubsection{4.8.4 The same pattern without the cue block, on the
100-patch
pilot}\label{the-same-pattern-without-the-cue-block-on-the-100-patch-pilot}

All six variants were issued once on a single 100-patch manifest. They
are weak evidence and are reported as a consistency check rather than an
estimate: n is 100, the manifest is class-stratified, and four of the
six sit at or near the 1/9 floor.

{\def\LTcaptype{none} 
\begin{longtable}[]{@{}
  >{\raggedright\arraybackslash}p{(\linewidth - 10\tabcolsep) * \real{0.1667}}
  >{\raggedright\arraybackslash}p{(\linewidth - 10\tabcolsep) * \real{0.1667}}
  >{\raggedright\arraybackslash}p{(\linewidth - 10\tabcolsep) * \real{0.1667}}
  >{\raggedright\arraybackslash}p{(\linewidth - 10\tabcolsep) * \real{0.1667}}
  >{\raggedright\arraybackslash}p{(\linewidth - 10\tabcolsep) * \real{0.1667}}
  >{\raggedright\arraybackslash}p{(\linewidth - 10\tabcolsep) * \real{0.1667}}@{}}
\toprule\noalign{}
\begin{minipage}[b]{\linewidth}\raggedright
100-patch pilot, n = 100
\end{minipage} & \begin{minipage}[b]{\linewidth}\raggedright
cue block
\end{minipage} & \begin{minipage}[b]{\linewidth}\raggedright
arity
\end{minipage} & \begin{minipage}[b]{\linewidth}\raggedright
closing paragraph
\end{minipage} & \begin{minipage}[b]{\linewidth}\raggedright
accuracy
\end{minipage} & \begin{minipage}[b]{\linewidth}\raggedright
modal share
\end{minipage} \\
\midrule\noalign{}
\endhead
\bottomrule\noalign{}
\endlastfoot
D1 & no & nine & forbids a chosen answer & 0.1111 & 0.9899 \\
D3 & no & nine & forbids it, and adds ``Do not choose a single answer''
& 0.1100 & 0.9800 \\
D2 & no & top-3 & forbids commentary and reasoning only & 0.2000 &
0.7800 \\
D6 & yes & nine & forbids a chosen answer & 0.1828 & 0.8280 \\
D5 & yes & top-3 & nothing & 0.4040 & 0.2222 \\
\end{longtable}
}

The degeneracy tracks the prohibition and not the cue block. Both
nine-class variants that forbid a chosen answer predict one class on 98
or 99 responses in 100 and sit at chance, with or without the cues; the
two variants that do not forbid one are the two that spread their
predictions.

Two clause comparisons look available and neither settles anything. D3
is D1 with an explicit \emph{``Do not choose a single answer''} added on
top of a closing paragraph that already forbids one; it moves accuracy
by -0.0011 and modal share by -0.0099, which is what one would expect if
the clause is saturated and also what one would expect at a floor. And
D1 against D2 appears to vary cardinality with a prohibition present on
both sides, but the two prohibitions are different sentences: D1 forbids
a chosen answer and D2 does not, because a ranked list's first entry
\emph{is} one. That pair varies cardinality and the chosen-answer clause
together, which is the confound this section exists to remove.

\subsubsection{4.8.5 Parsing the arm that writes
prose}\label{parsing-the-arm-that-writes-prose}

The parser did not read D4's reasoning instead of its numbers. The line
grammar requires a line to end at its probability, so a sentence of
analysis never matches, and every one of the 1,552 and 1,553 scored D4
responses matched exactly nine class lines. Where a prose line did
match, it produced a duplicate class and the response was failed rather
than scored: 110 of the 248 validation failures are the model emitting
the nine-line block twice.

\begin{center}\rule{0.5\linewidth}{0.5pt}\end{center}

\section{5. Discussion}\label{discussion}

\subsection{5.1 What the format effect is not, and which sentence it
is}\label{what-the-format-effect-is-not-and-which-sentence-it-is}

Eight explanations are available for a fourteen-point accuracy drop. The
design rules out seven, and the eighth is the answer.

It is not the cue block: both variants carry the same nine-line
per-class description, verbatim, and the difference in section 4.1 is
measured with that held fixed. It is not output length or formatting
burden as such, because if the cost scaled with how much structured text
the model has to emit, the ranked top-3 would sit between the single
label and the nine-class distribution, and almost the entire accuracy
cost falls in the one step between those two prompts. The variant that
isolates the prohibition settles this from the other side: it emits
about thirteen times as many completion tokens as the ranked top-3 and
matches its accuracy.

It is not constrained decoding, because we never constrain decoding. We
ask, and the model complies or does not. Lee et al.~(2026) put the same
question to correctness with the decoder available and held out as a
factor, and located most of the loss in the instruction itself (section
2.1). Ours never had a decoder to constrain, so two designs whose
confounds cannot overlap agree on where the cost enters.

It is not truncation. Every response in both formats at both centers
terminated at the model's own stop rather than at the token ceiling, 0
of 34,368 on two agreeing indicators, and D6's longest response anywhere
is 87 tokens against a budget of 256. D4, which makes the same
nine-class request without the prohibition, does truncate at that
budget, which is why section 3.12 crosses the budget with the prompt.

Nor is it a parse artefact, which the data rules out most cleanly:
counting unparseable responses as errors and comparing both variants on
the whole split makes the effect grow rather than shrink, so excluding
the failures was costing us about three points. Nor is the asymmetric
parse criterion doing the work, since the two formats fail for disjoint
reasons (section 4.2).

And it is not the number of classes. Section 4.8 holds the cue block,
the model, the decoding, the requested cardinality and the token budget
fixed and removes only the sentence forbidding commentary, reasoning and
a chosen answer; accuracy rises by about nineteen points at both
centers. Holding the absence of that sentence fixed and varying the
cardinality instead moves accuracy by an amount containing zero at both.
The prohibition carries the effect, and the cardinality carries none of
it that this design can detect.

Two facts from section 4.8 fix the interpretation. Given four times the
token budget the prohibited variant emits the same number of completion
tokens and returns the same accuracy, so it is not running out of room;
it has been told not to use it. And what it produces is not a degraded
classifier but one that has stopped choosing, putting four fifths of its
scored validation predictions on a single class and never using two of
the nine, against a fifth for the same request without the prohibition.

So the treatment is a sentence practitioners write on purpose. Some
version of ``return only the numbers, no commentary, no reasoning, no
chosen answer'' is what one adds to get parseable output out of a chat
model, and on this model and this task it is what destroys the
classifier. What we cannot say is which of its three clauses does the
work, because the prohibition forbids all three at once and the control
removes all three at once. The token counts point at the reasoning
clause, 480 against 86 when the model is allowed to deliberate.
Separating the three needs two more variants, and if only one were run
it should be the one that permits reasoning while still forbidding a
chosen answer, because that is the arm the evidence already leans on and
the one that can falsify it. If accuracy recovers with a chosen answer
still forbidden, the reasoning clause carries the effect; if it does
not, the reverse.

Two consequences do not follow. A nine-class elicitation does not become
the thing to reach for, since section 4.8 prices the repair at thirteen
times the completion tokens of a ranked top-3 and about five points of
coverage. And no measured value changes: what the control revises is the
sentence saying what the difference is a difference \emph{between}.

\subsection{5.2 What the failure is, and what triggers
it}\label{what-the-failure-is-and-what-triggers-it}

D6's characteristic failure is a vector of nine zeros rather than a
malformed answer, and the two blind-image controls of section 4.7 narrow
what it can be attributed to.

The first excludes a language-prior account. If the nine-class demand
made the model stop consulting the image and generate a plausible
nine-vector from language priors, removing the image would produce that
vector; instead D6 produces zeros, and the format that falls back on a
language prior is the top-3 one. Uninformative input in general does not
cause the failure either, since seeded noise restores a normal parse
rate. The trigger is a \emph{featureless} field, which is the property
the two classes it fails on share: adipose is described in the cue block
as large empty rounded cells and background as white or near-empty
glass, and they are the two least-textured classes in the dataset. On
7.6 and 7.7 percent of real patches, D6 behaves as though the patch were
blank.

The second adds the prohibition as a necessary condition: neither a
featureless image nor the prohibition produces the refusal alone. The
sentence that costs the accuracy is therefore the same one that, on a
featureless field, converts the model from answering badly to not
answering at all. Three things this paper reports separately have one
cause: the accuracy collapse, the degeneracy of the predictions, and the
refusal on featureless input.

What that sentence does inside the model is still unknown. Naming it is
not the same as knowing the mechanism, and the deployment setting that
makes the question worth asking is exactly the one that withholds the
weights needed to answer it. An enumeration account, under which a
nine-way demand consumes some limited integration capacity, is ruled out
by section 4.8, because the enumeration demand is free. If a capacity
mechanism is involved at all, what consumes it is the instruction to
withhold reasoning and an answer, which is a different and more testable
claim.

\subsection{5.3 Whether the cause is the format or the
capacity}\label{whether-the-cause-is-the-format-or-the-capacity}

Fan's capacity account (section 2.1) is the strongest challenge to how
this paper names its result, and section 4.6 answers it with a
measurement. It has two mechanisms, and neither survives here.

Truncation is a property of a run rather than of a theory, so we checked
ours: the truncation rate across all 34,368 responses in the four sealed
full-split runs is 0.000000 {[}0.000000, 0.000112{]} on two agreeing
indicators, and the nine-class format uses about a third of the budget
it is given. Our cue block is also byte-identical across the two
variants, which is the confound Fan's information-matched prose controls
are constructed to remove; we get it by construction rather than by
matching.

Fan's second mechanism needs capacity to be varied, and the ladder of
section 4.6 varies it across four models at both centers. The result
does not follow the capacity account's ordering: the cost is a flat band
and one outlier rather than a monotone function of headroom, and no
reading of capacity accounts for two models matched on task accuracy
differing in cost by a factor of five to seven. What the ladder does not
do is identify what the cost \emph{is} a function of, for the reasons
section 4.6 gives.

\subsection{5.4 Non-monotone routing value, and what a practitioner
should take from
this}\label{non-monotone-routing-value-and-what-a-practitioner-should-take-from-this}

The accuracy story and the routing story come apart. Ordered by how much
the format asks for, accuracy falls at every step and almost all of the
fall is in the last one; routing value instead is at chance at one
number, AUROC 0.4975, carries a real ordering at three, +0.0493, and is
back at chance at nine on the patches it can score, +0.0020.

The natural reading is that a usable confidence ordering needs the model
to be doing the task and to have somewhere to put its uncertainty. One
number gives it nowhere, and the result is a four-valued signal that
ties 91 percent of pairs. Three numbers plus a residual is enough room
to express doubt and little enough demand to leave the classification
intact. Nine numbers \emph{under the prohibition} stop it doing the
task. The upper edge of that window is not where we originally placed it
and we have not measured where it is: nine-without-the-prohibition is a
fourth rung whose routing value we did not measure, because the run was
issued to settle the causal question and the routing endpoints were not
part of that declaration. Whether the window has an upper edge at all,
or only a prohibition cliff, is open.

One thing here we cannot explain. The elicited object's ranking
efficiency more than doubles between centers, 0.1394 to 0.3851, with
disjoint intervals, while the first-token signal read from the same
calls sits at 0.5026 and 0.5059. Section 4.5 excludes the two candidate
explanations: not score resolution, since the difference survives
quantization down to two bins, and not a property of the patches or the
centers, since a signal read from the same responses is stable across
them. It matters for deployment more than most of what we report,
because it says the quality of an elicited confidence ordering is a
property of the site as well as of the model, and a triage threshold
tuned at one center may not mean the same thing at another.

Four recommendations follow, in order of how well the evidence supports
them.

\textbf{Do not tell a classifier not to reason and not to choose.} The
instruction that costs about fourteen accuracy points here is not the
request for nine numbers; it is the closing sentence, and removing it
while leaving everything else alone recovers essentially all of the
cost. The sentence is not an accident, it is what one writes to make a
chat model emit parseable output, and it is in a great many working
prompts for that reason. If a classification prompt of yours carries
one, measure what it costs before keeping it. The cost is not uniform
across models, so measure it on yours.

\textbf{Still ask for a short ranked list rather than a distribution.}
Once neither prompt suppresses an answer the two formats are not
distinguishable in accuracy on this task. What separates them is price:
a ranked top-3 costs 36.6 completion tokens per call and complies 99
percent of the time, while a nine-class distribution allowed to reason
first costs about 480 and complies 86 percent of the time. The ranked
list is the better buy, not the more accurate object.

\textbf{Read the first-token logits if your endpoint exposes them.} They
are strictly better here than anything we elicited, they cost nothing,
and they are stable across centers where the elicited object is not.

\textbf{If your endpoint returns text only, a single elicited ranked
top-3 is defensible.} It beats the stated scalar at both centers by
+0.0244 and +0.1668, and is not distinguishable from a five-query
paraphrase-agreement baseline costing 5.79 times as much. We cannot tell
you it matches that baseline, because this dataset cannot resolve a
difference of practical size at any sample size.

\subsection{5.5 Two limits on what a reproducibility apparatus can
see}\label{two-limits-on-what-a-reproducibility-apparatus-can-see}

Three of this paper's own results were replaced by checks run on them,
and the shape of the three failures bears on how work of this kind
should be reported. Two were failures of what a quantity was a property
of rather than of how it was computed. The elicited object's accuracy
advantage over the stated scalar was proposed as a secondary headline on
class-balanced subset values around +0.049, and at natural prevalence it
reverses sign on validation, -0.0264 (section 4.3). The five-query
comparator it was measured against extracted its labels with a regex
sweep over the whole response, which overrides the model's own stated
label on 1.93 percent of full-split responses; re-parsing removed an
external-center superiority (section 4.4.1). In both cases the artifact
reproduces exactly and re-running it would confirm the same number,
because what was wrong was not the computation: in the first the number
described the wrong population, and in the second the comparator was
labelled by a different rule from the object it was compared against.

The third was a failure of design description. The two prompts the
headline compares were described as differing only in the requested
response format, and they differ in two ways (section 3.3). Here the
quantity was right as well: the runs reproduce, the difference is real,
the interval is correct, and what was wrong was the noun naming what the
difference is \emph{between}. A design that varies two things and names
one of them passes every check a results log can run, because nothing in
a results log reads the prompt. Two practices follow. A re-parse under
an independent extractor deserves the same standing as a re-run. And a
paper whose treatment is a prompt should diff its prompts, print the
diff, and account for every line of it, because the difference between
two prompts \emph{is} the experiment.

\begin{center}\rule{0.5\linewidth}{0.5pt}\end{center}

\section{6. Limitations}\label{limitations}

Ordered by how much each costs the paper, and each stated with the
measurement that establishes it.

\subsection{6.1 Generality, and one serving
path}\label{generality-and-one-serving-path}

Everything here is nine-class colorectal tissue classification on one
benchmark. Nothing establishes that a format effect of this size exists
on another task, another label space, or another modality, and an
elicitation-format result ages with the model generation.

The routing half does not replicate on a second model. On
Gemma-3-27b-it, the non-medical base model of the same lineage, run on
the same 1,800 patches with the same prompt and decoding, the elicited
top-3 entropy's gap to its own recomputed floor is +0.0183 {[}-0.0194,
+0.0396{]}, containing zero; max probability and margin contain zero as
well. The format is not the problem, as Gemma-3's parse rate is 1.0000
{[}0.9979, 1.0000{]}, better than MedGemma's. Nor is it that these
patches carry no routing signal, because the first-token logits from the
same calls clear the floor by +0.0637 {[}+0.0267, +0.0892{]}. The
elicited object specifically fails, we cannot say why, and no claim of
generality across models is available for the routing result.

The capacity ladder answers less than it appears to: section 4.6 rules
out one explanation without supplying another, for the reasons given
there. A ladder that varied capacity within one lineage at fixed
post-training remains the experiment that would settle it, and the
ladder covers non-reasoning models only.

Every call in this study, across all four models and every variant, went
through the single hosted endpoint of section 3.2, so the nine-zeros
response cannot be separated from this deployment of MedGemma-27b-it.
Unlike the resolution limit of section 6.3 this bound is reachable:
re-running D5 and D6 on the two 1,800-patch subsets against a second
serving path for the same published checkpoint costs 7,200 calls.

``Cross-center'' is also a weaker shift than the phrase implies, for the
reasons given in section 3.1.

\subsection{6.2 A better routing score on these patches may buy nothing
clinically}\label{a-better-routing-score-on-these-patches-may-buy-nothing-clinically}

Pathology is a domain where base-model competence sets a low ceiling,
and our own per-class results make that concession sharper. The model
sits at 0.0100 {[}0.0090, 0.0110{]} selective accuracy on mucus and
0.0012 {[}0.0010, 0.0015{]} on cancer-associated stroma at the
validation center, against 0.5988 {[}0.5916, 0.6061{]} on adenocarcinoma
epithelium. For the two weak classes there is no routing story at all:
nothing to route because nothing is right. The aggregate also hides a
signal that actively misorders. The per-class gap to a within-class
random ordering is +0.2615 for adenocarcinoma epithelium, +0.1381 for
adipose and +0.1322 for normal colon mucosa, against -0.2857 for debris,
-0.0781 for smooth muscle and -0.0694 for lymphocytes, with the pattern
repeating at the external center. An area that is positive overall while
the signal misorders three of nine classes is a weaker object than the
aggregate suggests, and this paper does not claim a deployable system.

\subsection{6.3 The comparator comparison is permanently underpowered,
and its patch sets are
biased}\label{the-comparator-comparison-is-permanently-underpowered-and-its-patch-sets-are-biased}

The two preregistered comparators are five-query signals whose traces
exist only on 1,800-patch subsets, and no full-scale version was ever
run. The realised minimum detectable effect is 0.0517 on validation and
0.0526 at the external center, against the 0.02 practical threshold, and
section 3.8 establishes that no sample available on this dataset reaches
it: the whole splits would deliver 0.0230 and 0.0286, and pooling is
foreclosed because the claim is per-center. The equivalence clause is
not undecided pending more data; there is no more data.

The subsets are also not representative of their splits, and in opposite
directions: subset minus rest is +0.0466 {[}+0.0108, +0.0814{]} on
validation and -0.0548 {[}-0.0920, -0.0184{]} at the external center,
both excluding zero.

What remains on that basis is this: both five-query comparators and the
whole preregistered twelve-test family; the cue-free variant and
therefore the cue-block effect in section 4.1; the confirmatory
matched-cardinality control in section 4.5; the capacity ladder and the
blind-image control; and the second-temperature check. The nine-class
collapse and its failure concentration are not on that list: both are
measured at natural prevalence on both whole splits, at 17,184 calls,
which moved argmax accuracy from 0.1825 to 0.2223 and 0.2720.

\subsection{6.4 Measurement conventions, reproducibility of the object,
and two reporting
choices}\label{measurement-conventions-reproducibility-of-the-object-and-two-reporting-choices}

Three conventions could each have gone the other way, and each is
reported in both columns. The extraction convention changes a
preregistered comparison: every comparator available to us was labelled
by a regex sweep over the whole response while the elicited object was
not, and correcting that removes the external-center superiority
(section 4.4.1). The correction is not free of judgement, since
reproducing the comparator's reference values requires the sweep-based
convention and the stated-label alternative gives a materially different
baseline, +0.0611 {[}+0.0395, +0.0812{]}. The coverage treatment changes
one verdict, the nine-class control's at both centers, and none for the
primary variant.

The elicited object is not reproducible across decodes. At temperature
0.7 on the same patches the entropy scores correlate with the greedy
ones at Spearman only +0.5770 {[}+0.5413, +0.6110{]} and the two runs
agree on 0.7660 {[}0.7456, 0.7853{]} of their argmax predictions. The
routing signal survives the change, with a gap to floor of +0.0621
{[}+0.0383, +0.1006{]}, but the paired difference against greedy is
-0.0269 {[}-0.0512, +0.0011{]}, containing zero by about a thousandth,
so a real degradation cannot be excluded, and the run cannot separate
temperature from sampling noise (section 3.10). A practitioner who
reruns the same prompt at a nonzero temperature gets a materially
different confidence ordering.

No calibration metric is reported. The endpoint is deferral, where what
matters is whether a score \emph{orders} errors, and a monotone
recalibration leaves every ranking metric here unchanged while moving
every calibration metric (Brier, 1950; Naeini et al., 2015; Guo et al.,
2017; Nixon et al., 2019). The nine-class format's characteristic output
is also a vector of nine zeros, which has no calibration to measure, so
the comparison would run on a different patch population from every
other comparison here. Separately, we report area under the
accuracy-coverage curve rather than the generalized risk-coverage
measure of Traub et al.~(2024), which changes metric rankings on five of
six benchmarks; the metric set was closed in the preregistration before
we retrieved that record, so adding it now would be an endpoint chosen
after the data.

\subsection{6.5 The resampling unit is the patch and it should be the
patient}\label{the-resampling-unit-is-the-patch-and-it-should-be-the-patient}

Patches from one slide are not independent, so a patch-level bootstrap
understates the standard error by roughly the square root of the design
effect. The released PathMNIST arrays carry images and labels and no
slide or patient identifier, so a clustered resample is unavailable at
any price. Every interval in this paper is narrower than a
patient-clustered interval would be. The same limitation applies to
every published result on this benchmark.

\begin{center}\rule{0.5\linewidth}{0.5pt}\end{center}

\section{7. Conclusion}\label{conclusion}

The sentence that asks a medical vision-language model for its answer is
not packaging. Holding a histopathology prompt fixed down to its
per-class cue block and changing only the requested response format,
asking MedGemma-27b-it for a probability over all nine classes rather
than a ranked top-3 costs about fourteen accuracy points at each of two
clinical centers and takes the classifier to roughly twice chance.

What in that change does it is one sentence rather than the number of
classes. Only the nine-class prompt closes by forbidding commentary,
reasoning and a chosen answer. A third variant making the same
nine-class demand without that sentence recovers essentially the whole
of the effect at matched cardinality and token budget, while varying the
cardinality with the sentence absent from both prompts moves accuracy by
an amount whose interval contains zero at both centers. The prohibition
carries the cost; the number of classes carries none we can detect, and
what it costs instead is coverage.

The prohibited variant's only usable confidence signal is its own
failure to produce one. It emits a vector of nine zeros on about one
patch in thirteen, those patches concentrate on the two least-textured
classes, and deferring them first accounts for the whole of its
measurable selective-accuracy gain. Two blind-image controls say what
that refusal is a property of, and the answer is the same sentence: a
blank image reproduces the nine-zeros response on every call while the
ranked top-3 format returns a well-formed prior from the same empty
input, seeded noise restores the nine-class format to a normal parse
rate, and the nine-class variant \emph{without} the prohibition never
produces the refusal at all. It needs both a featureless image and the
instruction not to answer. Across four vision-capable models the cost is
separately not a smooth function of a model's accuracy on the task, so
it is not headroom under another name.

The ranked top-3 does route, and beats the single stated confidence it
replaces at both centers. It does not beat a five-query
paraphrase-agreement baseline, and this dataset cannot resolve a
difference of practical size between them at any sample size, so that
comparison is closed rather than deferred. And it does not beat the
entropy of the model's own first-token logits, which arrive free in the
same response and are stable across centers where the elicited object is
not. Read the logits where the endpoint exposes them; where it does not,
ask for a short ranked list, which is the cheapest of the formats that
work rather than the most accurate; and do not tell a classifier not to
reason and not to choose.

What remains open is one level below where we left it. We can rule out
the cue block, output length, constrained decoding, truncation, parse
treatment, the parse criterion and the requested cardinality, and we can
name the sentence that does the work. We cannot say which of its three
clauses does it, because the prohibition forbids commentary, reasoning
and a chosen answer together and the control removes all three together.
The token counts point at the reasoning clause, and the experiment that
would settle it permits reasoning while still forbidding a chosen
answer. That is a prompt, not a program.

\begin{center}\rule{0.5\linewidth}{0.5pt}\end{center}

\section{Declarations}\label{declarations}

\subsection{Ethics approval and consent to
participate}\label{ethics-approval-and-consent-to-participate}

Not applicable. This study used public, de-identified datasets and
collected no new patient data; no human participants were enrolled, and
no ethics approval or consent was required.

\subsection{Consent for publication}\label{consent-for-publication}

Not applicable.

\subsection{Data availability}\label{data-availability}

Every dataset used here is public. PathMNIST is distributed as part of
MedMNIST v2 (Yang et al., 2023); we use the 224-pixel release. Its
source images are NCT-CRC-HE-100K and CRC-VAL-HE-7K (Kather et al.,
2018), deposited on Zenodo under DOI 10.5281/zenodo.1214456. No new
patient data was collected, no human subjects were involved, and no
ethics approval was required or sought. The split definitions, patch
manifests and per-patch labels are the released ones; the 544-patch
extension drawn for this study is released with its seed and its
selection rule.

\subsection{Code availability}\label{code-availability}

All code, configuration, prompts and analysis needed to reproduce every
number, table and figure in this paper are released at
https://github.com/anayy09/elicitation-format-medvlm. The release that
produced the results reported here is version 2.0.0, archived under DOI
10.5281/zenodo.22667810; DOI 10.5281/zenodo.22667809 resolves to the
most recent version. The repository contains the inference client, the
parser, the selective-prediction and bootstrap implementations, one
configuration file per run, the prompt files verbatim, the analysis
scripts that emit each reported quantity, the figure code, the test
suite, and instructions for running the pipeline from a fresh
environment against any OpenAI-compatible endpoint.

Each inference run is released as a sealed directory containing the
exact configuration used, its SHA-256 hash, the patch manifest, the raw
model responses, the per-patch token usage, the run metadata, and a
checksum manifest over all of it, re-verified at analysis time. Because
the served endpoint is not itself reproducible by a third party (section
3.2), releasing the raw responses is what makes every analysis
downstream of inference reproducible without re-running any inference.
The preregistration is released alongside, together with an append-only
log of every computed quantity, and every number in this manuscript is
attributed to exactly one entry of that log through Supplementary Table
S1.

\subsection{Acknowledgments and computational
resources}\label{acknowledgments-and-computational-resources}

All model inference for this study was performed on the University of
Florida's NaviGator AI service, operated by UF Research Computing, which
provided the OpenAI-compatible endpoint and the GPU capacity behind it.
We thank UF Research Computing for providing and maintaining the
service.

\subsection{Funding}\label{funding}

No external funding supported this work. Computational resources were
provided by the University of Florida as described above.

\subsection{Competing interests}\label{competing-interests}

The authors declare no competing interests.

\subsection{Author contributions}\label{author-contributions}

A.S. conceived the study, designed the experiments, implemented the
evaluation pipeline and all models, conducted the analysis, and wrote
the manuscript. Am.S. and Ar.S. provided supervision, methodological
guidance, and critical review of the manuscript. All authors read and
approved the final manuscript.

\begin{center}\rule{0.5\linewidth}{0.5pt}\end{center}

\section{References}\label{references}

Where a peer-reviewed version of record exists, it is what is cited. Six
of the thirty-eight entries are preprints with no peer-reviewed version
at the time of writing and are marked \textbf{{[}preprint{]}}.

Aggarwal, P., Madaan, A., Yang, Y., Mausam (2023). Let's Sample Step by
Step: Adaptive-Consistency for Efficient Reasoning and Coding with LLMs.
\emph{Proceedings of EMNLP 2023}, 12375-12396. DOI
10.18653/v1/2023.emnlp-main.761.

Begoli, E., Bhattacharya, T., Kusnezov, D. (2019). The need for
uncertainty quantification in machine-assisted medical decision making.
\emph{Nature Machine Intelligence} 1(1), 20-23. DOI
10.1038/s42256-018-0004-1.

Brier, G. W. (1950). Verification of Forecasts Expressed in Terms of
Probability. \emph{Monthly Weather Review} 78(1), 1-3. DOI
10.1175/1520-0493(1950)078\textless0001:VOFEIT\textgreater2.0.CO;2.

Chow, C. K. (1970). On optimum recognition error and reject tradeoff.
\emph{IEEE Transactions on Information Theory} 16(1), 41-46. DOI
10.1109/TIT.1970.1054406.

Dolezal, J. M., Srisuwananukorn, A., Karpeyev, D., Ramesh, S., Kochanny,
S., Cody, B., et al. (2022). Uncertainty-informed deep learning models
enable high-confidence predictions for digital histopathology.
\emph{Nature Communications} 13(1). DOI 10.1038/s41467-022-34025-x.

El-Yaniv, R., Wiener, Y. (2010). On the Foundations of Noise-free
Selective Classification. \emph{Journal of Machine Learning Research}
11(53), 1605-1641. https://jmlr.org/papers/v11/el-yaniv10a.html

Fan, H. (2026). Capacity, Not Format: Rethinking Structured Reasoning
Failures. arXiv:2606.09410. \textbf{{[}preprint{]}}

Geifman, Y., El-Yaniv, R. (2017). Selective Classification for Deep
Neural Networks. \emph{Advances in Neural Information Processing Systems
30}, 4878-4887. arXiv:1705.08500.

Gemma Team (2025). Gemma 3 Technical Report. arXiv:2503.19786.
\textbf{{[}preprint{]}}

Groot, T., Valdenegro-Toro, M. (2024). Overconfidence is Key: Verbalized
Uncertainty Evaluation in Large Language and Vision-Language Models.
\emph{Proceedings of TrustNLP, the Fourth Workshop on Trustworthy
Natural Language Processing, at NAACL 2024}. arXiv:2405.02917.

Guo, C., Pleiss, G., Sun, Y., Weinberger, K. Q. (2017). On Calibration
of Modern Neural Networks. \emph{Proceedings of the 34th International
Conference on Machine Learning}, PMLR 70, 1321-1330. arXiv:1706.04599.

Huang, Z., Bianchi, F., Yuksekgonul, M., Montine, T. J., Zou, J. (2023).
A visual-language foundation model for pathology image analysis using
medical Twitter. \emph{Nature Medicine} 29(9), 2307-2316. DOI
10.1038/s41591-023-02504-3.

Kadavath, S., Conerly, T., Askell, A., Henighan, T., Drain, D., Perez,
E., et al.~(2022). Language Models (Mostly) Know What They Know.
arXiv:2207.05221. \textbf{{[}preprint{]}}

Kamath, A., Jia, R., Liang, P. (2020). Selective Question Answering
under Domain Shift. \emph{Proceedings of the 58th Annual Meeting of the
Association for Computational Linguistics}, 5684-5696. DOI
10.18653/v1/2020.acl-main.503.

Kather, J. N., Halama, N., Marx, A. (2018). 100,000 histological images
of human colorectal cancer and healthy tissue. Zenodo. DOI
10.5281/zenodo.1214456.

Kather, J. N., Pearson, A. T., Halama, N., Jaeger, D., Krause, J.,
Loosen, S. H., et al.~(2019). Deep learning can predict microsatellite
instability directly from histology in gastrointestinal cancer.
\emph{Nature Medicine} 25(7), 1054-1056. DOI 10.1038/s41591-019-0462-y.

Kompa, B., Snoek, J., Beam, A. L. (2021). Second opinion needed:
communicating uncertainty in medical machine learning. \emph{npj Digital
Medicine} 4(1). DOI 10.1038/s41746-020-00367-3.

Lee, I. Y., D'Antoni, L., Berg-Kirkpatrick, T. (2026). The Format Tax.
arXiv:2604.03616. \textbf{{[}preprint{]}}

Leibig, C., Allken, V., Ayhan, M. S., Berens, P., Wahl, S. (2017).
Leveraging uncertainty information from deep neural networks for disease
detection. \emph{Scientific Reports} 7(1). DOI
10.1038/s41598-017-17876-z.

Lin, S., Hilton, J., Evans, O. (2022). Teaching Models to Express Their
Uncertainty in Words. \emph{Transactions on Machine Learning Research}.
arXiv:2205.14334.

Long, D. X., Nguyen, N. H., Sim, T., Dao, H., Joty, S., Kawaguchi, K.,
Chen, N. F., Kan, M.-Y. (2025). LLMs Are Biased Towards Output Formats!
Systematically Evaluating and Mitigating Output Format Bias of LLMs.
\emph{Proceedings of NAACL 2025 (Volume 1: Long Papers)}, 299-330. DOI
10.18653/v1/2025.naacl-long.15.

Lu, M. Y., Chen, B., Williamson, D. F. K., Chen, R. J., Liang, I., Ding,
T., et al.~(2024). A visual-language foundation model for computational
pathology. \emph{Nature Medicine} 30(3), 863-874. DOI
10.1038/s41591-024-02856-4.

Moor, M., Banerjee, O., Abad, Z. S. H., Krumholz, H. M., Leskovec, J.,
Topol, E. J., Rajpurkar, P. (2023). Foundation models for generalist
medical artificial intelligence. \emph{Nature} 616(7956), 259-265. DOI
10.1038/s41586-023-05881-4.

Naeini, M. P., Cooper, G., Hauskrecht, M. (2015). Obtaining Well
Calibrated Probabilities Using Bayesian Binning. \emph{Proceedings of
the AAAI Conference on Artificial Intelligence} 29(1), 2901-2907. DOI
10.1609/aaai.v29i1.9602.

Niazi, M. K. K., Parwani, A. V., Gurcan, M. N. (2019). Digital pathology
and artificial intelligence. \emph{The Lancet Oncology} 20(5),
e253-e261. DOI 10.1016/S1470-2045(19)30154-8.

Nixon, J., Dusenberry, M. W., Zhang, L., Jerfel, G., Tran, D. (2019).
Measuring Calibration in Deep Learning. \emph{CVPR 2019 Workshops},
38-41. arXiv:1904.01685.

Sclar, M., Choi, Y., Tsvetkov, Y., Suhr, A. (2024). Quantifying Language
Models' Sensitivity to Spurious Features in Prompt Design, or: How I
learned to start worrying about prompt formatting. \emph{International
Conference on Learning Representations (ICLR) 2024}. arXiv:2310.11324.

Sellergren, A., Kazemzadeh, S., Jaroensri, T., Kiraly, A., Traverse, M.,
Kohlberger, T., et al. (2025). MedGemma Technical Report.
arXiv:2507.05201. \textbf{{[}preprint{]}}

Tam, Z. R., Wu, C.-K., Tsai, Y.-L., Lin, C.-Y., Lee, H., Chen, Y.-N.
(2024). Let Me Speak Freely? A Study On The Impact Of Format
Restrictions On Large Language Model Performance. \emph{Proceedings of
EMNLP 2024: Industry Track}, 1218-1236. DOI
10.18653/v1/2024.emnlp-industry.91.

Taveekitworachai, P., Pongjirapat, N., Chaisutyakorn, K., Ittichaiwong,
P., Saengja, T., Pipatanakul, K. (2026). On the Robustness of Answer
Formats in Medical Reasoning Models. arXiv:2509.20866v2, 5 January 2026.
\textbf{{[}preprint{]}}

Tian, K., Mitchell, E., Zhou, A., Sharma, A., Rafailov, R., Yao, H.,
Finn, C., Manning, C. D. (2023). Just Ask for Calibration: Strategies
for Eliciting Calibrated Confidence Scores from Language Models
Fine-Tuned with Human Feedback. \emph{Proceedings of EMNLP 2023},
5433-5442. DOI 10.18653/v1/2023.emnlp-main.330.

Traub, J., Bungert, T. J., Lueth, C. T., Baumgartner, M., Maier-Hein, K.
H., Maier-Hein, L., Jaeger, P. F. (2024). Overcoming Common Flaws in the
Evaluation of Selective Classification Systems. \emph{Advances in Neural
Information Processing Systems 37}. arXiv:2407.01032.

van der Laak, J., Litjens, G., Ciompi, F. (2021). Deep learning in
histopathology: the path to the clinic. \emph{Nature Medicine} 27(5),
775-784. DOI 10.1038/s41591-021-01343-4.

Wang, X., Wei, J., Schuurmans, D., Le, Q., Chi, E., Narang, S.,
Chowdhery, A., Zhou, D. (2023). Self-Consistency Improves Chain of
Thought Reasoning in Language Models. \emph{International Conference on
Learning Representations (ICLR) 2023}. arXiv:2203.11171.

Wilson, E. B. (1927). Probable Inference, the Law of Succession, and
Statistical Inference. \emph{Journal of the American Statistical
Association} 22(158), 209-212. DOI 10.1080/01621459.1927.10502953.

Xiong, M., Hu, Z., Lu, X., Li, Y., Fu, J., He, J., Hooi, B. (2024). Can
LLMs Express Their Uncertainty? An Empirical Evaluation of Confidence
Elicitation in LLMs. \emph{International Conference on Learning
Representations (ICLR) 2024}. arXiv:2306.13063.

Xu, H., Usuyama, N., Bagga, J., Zhang, S., Rao, R., Naumann, T., et
al.~(2024). A whole-slide foundation model for digital pathology from
real-world data. \emph{Nature} 630(8015), 181-188. DOI
10.1038/s41586-024-07441-w.

Yang, J., Shi, R., Wei, D., Liu, Z., Zhao, L., Ke, B., Pfister, H., Ni,
B. (2023). MedMNIST v2: A Large-Scale Lightweight Benchmark for 2D and
3D Biomedical Image Classification. \emph{Scientific Data} 10(1), 41.
DOI 10.1038/s41597-022-01721-8.

\begin{center}\rule{0.5\linewidth}{0.5pt}\end{center}

\section{Figures}\label{figures}

Six figures, supplied as separate PDFs at both single-column (89 mm) and
double-column (183 mm) widths with 600 DPI PNG previews. Captions are
reproduced below in the order the text cites them.

Figures are numbered by order of first citation, so the two schematics
Methods cites are Figures 1 and 2 and the four data figures Results
cites are Figures 3 to 6. Each filename carries the number of the figure
it holds.

\begin{figure}
\centering
\includegraphics[width=1\linewidth,height=\textheight,keepaspectratio,alt={Figure 1. The inference pipeline, and where the manipulated variable sits in it. One patch from either clinical center enters one prompt whose only varying element is the sentence describing the requested response format. The prompt goes to a hosted model treated as a black box, and the returned text is parsed under a rule stated separately for each format. A single call yields three objects: a probability vector, a parse failure, and, where the endpoint returns logprobs, the top-5 logprobs at the first content-bearing token. Everything is compared against the same patches under the three baselines shown, each signal against its own recomputed random floor.}]{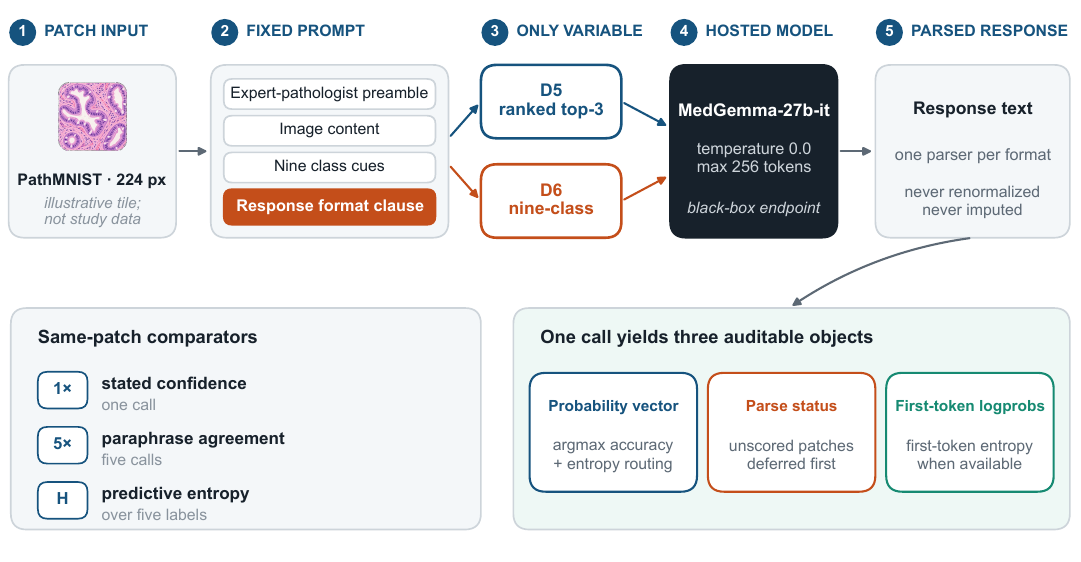}
\caption{\textbf{Figure 1. The inference pipeline, and where the
manipulated variable sits in it.} One patch from either clinical center
enters one prompt whose only varying element is the sentence describing
the requested response format. The prompt goes to a hosted model treated
as a black box, and the returned text is parsed under a rule stated
separately for each format. A single call yields three objects: a
probability vector, a parse failure, and, where the endpoint returns
logprobs, the top-5 logprobs at the first content-bearing token.
Everything is compared against the same patches under the three
baselines shown, each signal against its own recomputed random floor.}
\end{figure}

\begin{figure}
\centering
\includegraphics[width=1\linewidth,height=\textheight,keepaspectratio,alt={Figure 2. The analysis workflow, and the control that bounds each comparison. Each sealed run is verified against its own checksum manifest before it is read, then scored under two accounting bases that are reported side by side for every signal. The four comparisons the paper makes are shown with the quantity each is measured against, and beneath each is the control run to exclude its most likely alternative explanation. Every quantity resolves to one entry of the append-only results log.}]{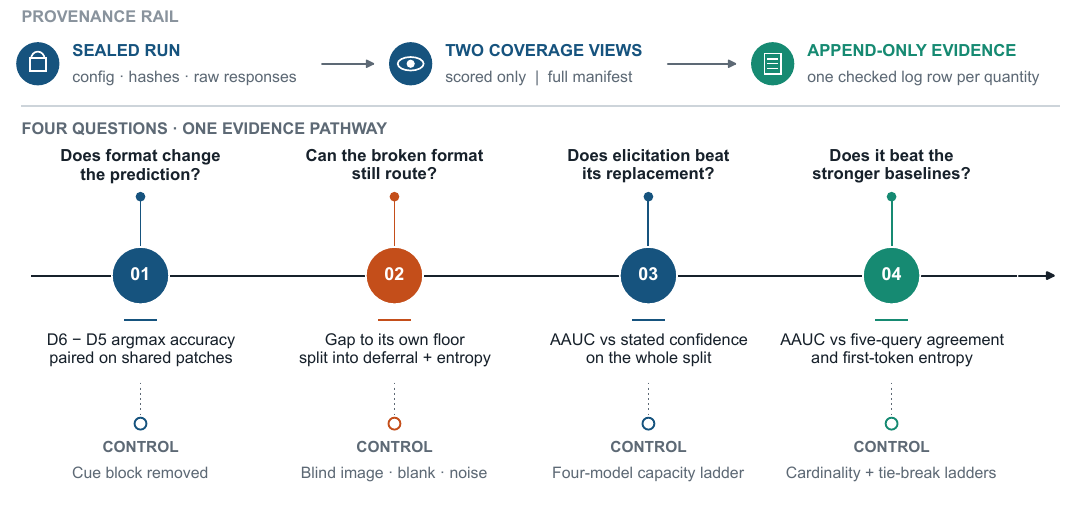}
\caption{\textbf{Figure 2. The analysis workflow, and the control that
bounds each comparison.} Each sealed run is verified against its own
checksum manifest before it is read, then scored under two accounting
bases that are reported side by side for every signal. The four
comparisons the paper makes are shown with the quantity each is measured
against, and beneath each is the control run to exclude its most likely
alternative explanation. Every quantity resolves to one entry of the
append-only results log.}
\end{figure}

\begin{figure}
\centering
\includegraphics[width=1\linewidth,height=\textheight,keepaspectratio,alt={Figure 3. What each confidence signal does to the accuracy-coverage curve. Selective accuracy against coverage at both clinical centers. Dotted horizontal lines are the random-ordering floors, which equal each system's own overall accuracy and therefore differ between signals. Curves are the mean over S = 200 tie-break seeds on a 200-point coverage grid. (a) validation, n = 10,004. (b) external test, n = 7,180.}]{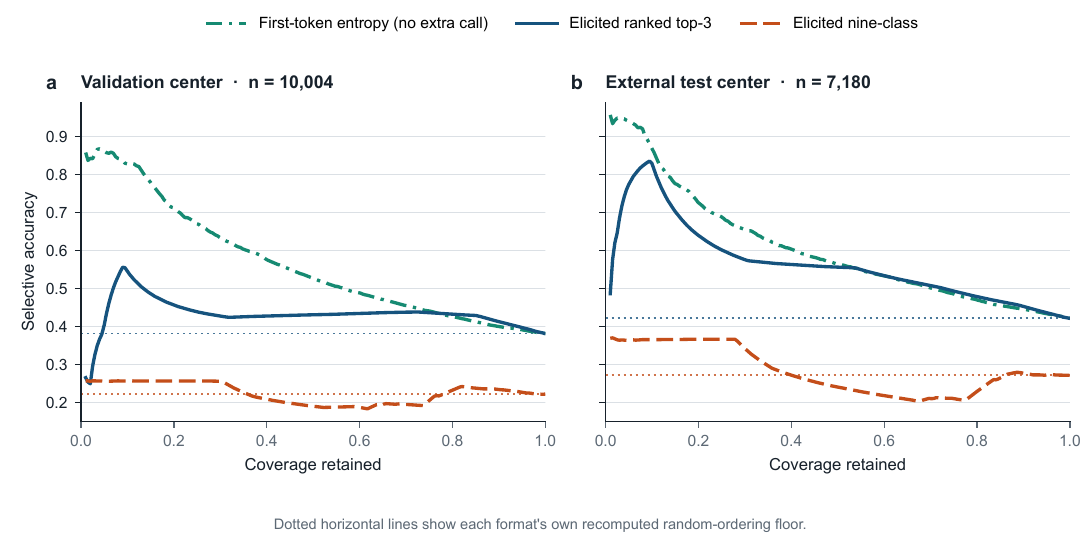}
\caption{\textbf{Figure 3. What each confidence signal does to the
accuracy-coverage curve.} Selective accuracy against coverage at both
clinical centers. Dotted horizontal lines are the random-ordering
floors, which equal each system's own overall accuracy and therefore
differ between signals. Curves are the mean over S = 200 tie-break seeds
on a 200-point coverage grid. (a) validation, n = 10,004. (b) external
test, n = 7,180.}
\end{figure}

\begin{figure}
\centering
\includegraphics[width=1\linewidth,height=\textheight,keepaspectratio,alt={Figure 4. The format effect. Paired patch-level differences with 95 percent paired-bootstrap intervals at B = 10,000, on the responses where both formats parsed: n = 9,163 on validation and n = 6,599 at the external center.}]{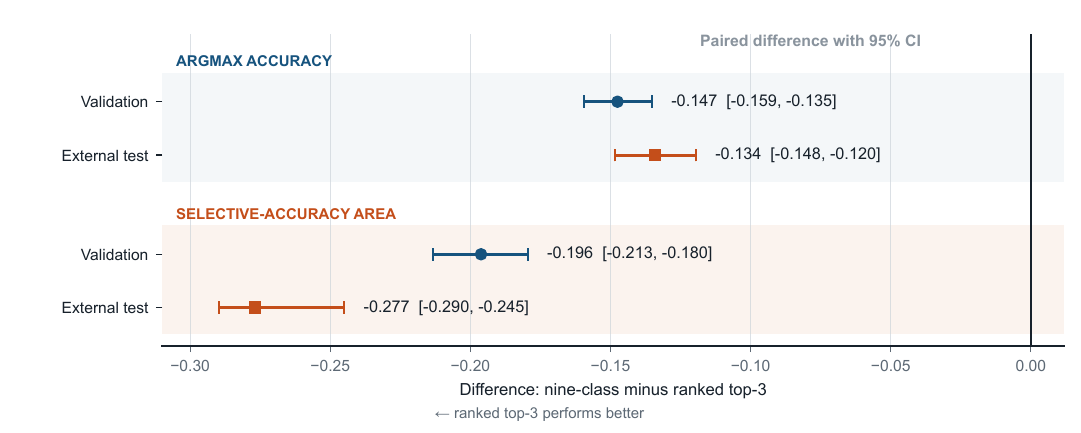}
\caption{\textbf{Figure 4. The format effect.} Paired patch-level
differences with 95 percent paired-bootstrap intervals at B = 10,000, on
the responses where both formats parsed: n = 9,163 on validation and n =
6,599 at the external center.}
\end{figure}

\begin{figure}
\centering
\includegraphics[width=1\linewidth,height=\textheight,keepaspectratio,alt={Figure 5. What the nine-class format's routing value is made of. Filled markers indicate an interval excluding zero, open markers one containing it. The shaded band is the preregistered practical threshold of 0.02; the dashed vertical line is the realised minimum detectable effect of the entropy row. (a) validation. (b) external test.}]{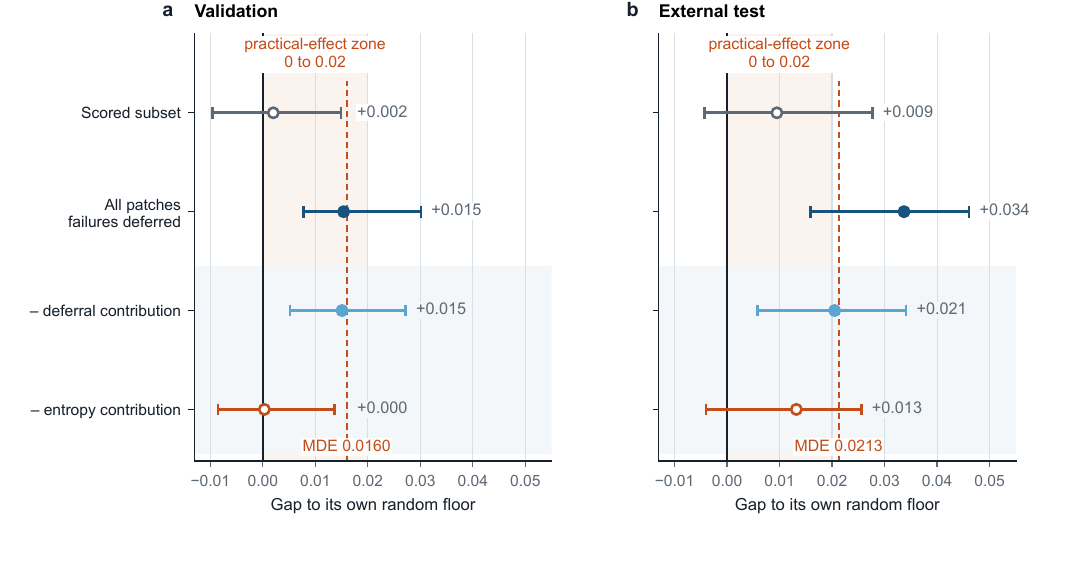}
\caption{\textbf{Figure 5. What the nine-class format's routing value is
made of.} Filled markers indicate an interval excluding zero, open
markers one containing it. The shaded band is the preregistered
practical threshold of 0.02; the dashed vertical line is the realised
minimum detectable effect of the entropy row. (a) validation. (b)
external test.}
\end{figure}

\begin{figure}
\centering
\includegraphics[width=1\linewidth,height=\textheight,keepaspectratio,alt={Figure 6. The routing league table. Differences in area under the accuracy-coverage curve with 95 percent paired-bootstrap intervals; filled markers exclude zero, open markers contain it. The first two groups are each signal against its own recomputed random floor and are not a head-to-head.}]{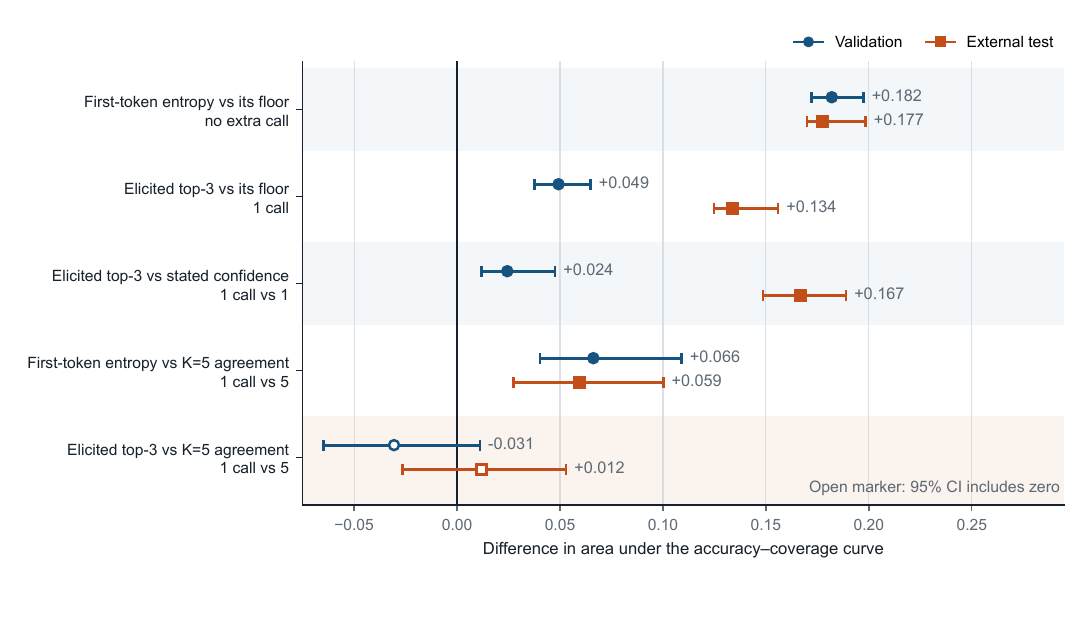}
\caption{\textbf{Figure 6. The routing league table.} Differences in
area under the accuracy-coverage curve with 95 percent paired-bootstrap
intervals; filled markers exclude zero, open markers contain it. The
first two groups are each signal against its own recomputed random floor
and are not a head-to-head.}
\end{figure}

\end{document}